\documentclass[10pt,twocolumn,letterpaper]{article}

\usepackage{iccv}
\usepackage{times}
\usepackage{epsfig}
\usepackage{graphicx}
\usepackage{amsmath}
\usepackage{amssymb}

\def\ie{\mbox{\textit{i.e.}, }}
\def\eg{\mbox{\textit{e.g.}, }}
\def\wrt{\mbox{\textit{w.r.t. }}}

\DeclareMathAlphabet\mathbfcal{OMS}{cmsy}{b}{n}

\def\0{{\bf 0}}
\usepackage{ dsfont }
\def\1{\mathds{1}}

\usepackage{enumitem}

\usepackage{graphicx}
\usepackage{color}
\usepackage{bbm}
\usepackage{bbding}
\usepackage{multirow}
\usepackage{algorithm}
\usepackage{algorithmic}
\usepackage{amsmath}
\usepackage{booktabs} 
\newcommand{\topline}{\toprule [0.1em]}
\newcommand{\midline}{\midrule [0.05em]}
\newcommand{\bottomline}{\bottomrule [0.1em]}
\usepackage{colortbl} 
\usepackage{xcolor}
\usepackage{xspace}
\usepackage{bigstrut}
\usepackage{dsfont}
\usepackage{pifont}
\usepackage{subcaption}

\usepackage{array}   
\definecolor{mygray}{rgb}{0.9,0.9,0.9}

\usepackage{caption} 
\usepackage[pagebackref=true,breaklinks=true,letterpaper=true,colorlinks,bookmarks=false]{hyperref}

\usepackage[capitalize]{cleveref}
\crefname{section}{Sec.}{Secs.}
\Crefname{section}{Section}{Sections}
\Crefname{table}{Table}{Tables}
\crefname{table}{Tab.}{Tabs.}

\iccvfinalcopy 

\ificcvfinal\fi

\begin{document}

\title{Learning Vision-and-Language Navigation 
                from YouTube Videos}

\author{
    Kunyang Lin\textsuperscript{\rm 1 2}\thanks{Equal contribution. Email: \{imkunyanglin, phchencs\}@gmail.com} ~~ 
    Peihao Chen\textsuperscript{\rm 1}\footnotemark[1] ~~ 
    Diwei Huang\textsuperscript{\rm 1} ~ 
    Thomas H. Li\textsuperscript{\rm 6} ~
    Mingkui Tan\textsuperscript{\rm 1 \rm 5}\thanks{Corresponding author. Email: mingkuitan@scut.edu.cn} ~
    Chuang Gan\textsuperscript{\rm 3 \rm 4} \\
    \textsuperscript{\scriptsize{\rm 1}}\small{South China University of Technology,}
    \textsuperscript{\scriptsize{\rm 2}}\small{Information Technology R\&D Innovation Center of Peking University,}\\
    \textsuperscript{\scriptsize{\rm 3}}\small{UMass Amherst,}
    \textsuperscript{\rm 4}\small{MIT-IBM Watson AI Lab,}
    \textsuperscript{\rm 5}\small{Key Laboratory of Big Data and Intelligent Robot, Ministry of Education,} \\
    \textsuperscript{\scriptsize{\rm 6}}\small{Peking University Shenzhen Graduate School} 
}

\maketitle
\ificcvfinal\thispagestyle{empty}\fi

\begin{abstract}

Vision-and-language navigation (VLN) requires an embodied agent to navigate in realistic 3D environments using natural language instructions. 
Existing VLN methods suffer from training on small-scale environments or unreasonable path-instruction datasets, limiting the generalization to unseen environments.
There are massive house tour videos on YouTube, providing abundant real navigation experiences and layout information. However, these videos have not been explored for VLN before.
In this paper, we propose to learn an agent from these videos by creating a large-scale dataset which comprises reasonable path-instruction pairs from house tour videos and pre-training the agent on it.
To achieve this, we have to tackle the challenges of automatically constructing path-instruction pairs and exploiting real layout knowledge from raw and unlabeled videos.
To address these, we first leverage an entropy-based method to construct the nodes of a path trajectory. Then, we propose an action-aware generator for generating instructions from unlabeled trajectories. Last, we devise a trajectory judgment pretext task to encourage the agent to mine the layout knowledge. Experimental results show that our method achieves state-of-the-art performance on two popular benchmarks (R2R and REVERIE). Code is available at \small{\url{https://github.com/JeremyLinky/YouTube-VLN}}
\end{abstract}
\vspace{-2em}

\section{Introduction}
\label{sec:intro}
An important goal of embodied artificial intelligence is to develop agents that can interact with humans in natural language to carry out real-world tasks.
Toward this goal, vision-and-language navigation (VLN)~\cite{VLN} is a rudimentary artificial intelligence task, requiring an indoor agent to navigate in unseen environments following natural instructions.
VLN has attracted widespread attention in the fields of computer vision and robotics due to its promising applications such as in-home robots~\cite{HomeRoboticsApplication} and warehouse assistants~\cite{liang2015automated}.

\begin{figure*}[t]
    \centering
	\includegraphics[width=1\linewidth]{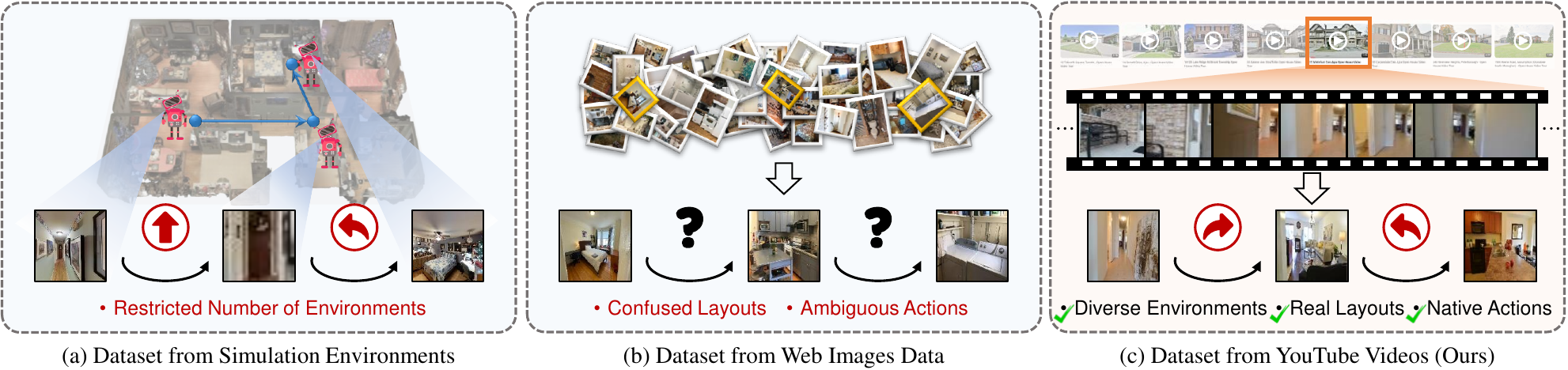}
	\caption{Comparison of different datasets for pre-training. Existing datasets are built from either \textbf{simulation environments}~(a) or \textbf{web images data}~(b). The former only covers limited environments, and the latter contains confusing layouts and ambiguous actions. Our dataset built from \textbf{YouTube videos}~(c) is able to provide diverse environments, real layouts and native actions simultaneously.
	}
    \label{fig:teaser}
\end{figure*}

One of the key challenges of VLN is the generalization ability of agents to unseen environments. 
Existing VLN methods attempt to cope with this challenge via self-supervised pre-training on vision-and-language datasets. As shown in Figure~\ref{fig:teaser}~(a), some previous works~\cite{Prompt, Transferable, Improving, AutoVLN} learn the agents on simulated navigation environments and manual-labeled data. The other works~\cite{Generic, Airbert, HOP} seek to construct path-instruction pairs by using web image data, which is shown in Figure~\ref{fig:teaser}~(b). Despite their promising performance, existing agents still suffer from the following limitations. 1) Training on simulated datasets is limited to a restricted number of environments. 2) Constructing a trajectory by simply concatenating web images leads to unreasonable room layouts, which hamper the agent to learn layout reasoning ability. As a result, VLN agents trained on such data are brittle to adapt to unseen environments.

Fortunately, there are massive house tour videos on YouTube, providing real navigation experiences and layout information but are still under-explored. We can be naturally inspired to let an agent learn VLN ability from such videos, thereby addressing the limitations of existing methods. An intuitive way is to model the navigation experiences as path-instruction pairs to train the agent.
Motivated by this, we propose a ``\textbf{Lily}'' agent who \textbf{L}\textit{earns V}\textbf{i}\textit{sion-and-}\textbf{L}\textit{anguage}\textit{ Navigation from }\textbf{Y}\textit{ouTube Videos}. Specifically, we first develop an in-domain pre-training dataset from house tour YouTube videos, namely \textbf{YouTube-VLN}, which comprises VLN-like path-instruction pairs. Our YouTube-VLN dataset has the advantages of diverse environments, real layouts, and native actions\footnote{The execution actions that objectively exist}, reducing the domain gap with VLN datasets, as illustrated in Figure~\ref{fig:teaser} (c). Then, we pre-train the agent using these path-instruction pairs. Benefiting from in-domain pre-training on our proposed dataset, our agent thus generalizes well to unseen environments.

Constructing and utilizing such a dataset, however, is still far from trivial work and remains an open problem due to the following challenges. 1) As the nodes in a trajectory are expected to be diverse and informative, it is hard to determine the locations of trajectory nodes from massive video frames and represent the visual content in a node. 2) Real VLN instructions include various action descriptions, but obtaining corresponding instructions from navigation clips is challenging due to the actions being implicit in videos. Thus it is nontrivial to acquire matching instruction on a trajectory. 3) Layout knowledge from real navigation experience is hard to mine and model, which impedes the agent of learning layout reasoning ability. 

In this paper, we address the above challenges as follows. To conquer challenge 1), we propose an entropy-based trajectory generation method. Specifically, we first envisage that the nodes of a trajectory should contain as many types of rooms as possible to diversify trajectories. Accordingly, we group the frames with the same room types in videos and consider each group as a node in the trajectory. Then, inspired by that low classification entropy image is reliable and contains rich information relevant to a specific class (room type in our case)~\cite{EATA}, the frame with the lowest classification entropy in a group is chosen to represent the visual content in a node.
To tackle challenge 2), we introduce an action-aware instruction generation method. Specifically, we adopt an action inverse model to pseudo-label the action along trajectories and fill them in the instructions via hand-designed rules.
To grapple with challenge 3), we devise a self-supervised pretext task. As we all know, humans often judge whether a navigation trajectory is reasonable based on the layout of the environment. Therefore, it is believed that an agent equipped with layout reasoning ability should be able to make similar judgment. To this end, we propose \textbf{trajectory judgment} pretext task to ask the agent to identify reasonable navigation trajectories, which further equips the model with the ability to reason environment layouts.

We empirically show that the diverse entropy-based  trajectory generation method and action-aware instruction generator allow us to harvest high-quality path-instruction pairs from YouTube house tour videos, resulting in the YouTube-VLN dataset. By integrating the self-supervised trajectory judgment task in pre-training a VLN agent, our Lily agent presents state-of-the-art performance on two mature and solid benchmarks (R2R~\cite{VLN}, REVERIE~\cite{REVERIE}). The proposed Lily agent reaches the first place on the R2R leaderboard in terms of success rate and outperforms the SOTA method under discriminative setting and generative setting by 2\% and 3\% \textit{w.r.t.} success rate, respectively.

Our main contributions are as follows:

$\bullet$ We unleash the huge potential of house tour videos for VLN. By leveraging these videos, we introduce a large-scale dataset containing real navigation path-instruction pairs for promoting VLN pre-training and a self-supervised pretext task for the learning of layout reasoning.

$\bullet$ Our diverse trajectory generation method, together with the action-aware instruction generator, creates informative and diverse trajectory nodes and produces matching instructions, both of which make the path-instruction pairs authentic and of high quality for training a VLN agent.

$\bullet$ The proposed trajectory judgment pretext task allows the model to build up an awareness of learning and reasoning the layout knowledge, which is crucial in the VLN task of indoor environments. We also empirically substantiate that the agent indeed learns the layout learning ability.

\section{Related Work}

\subsection{Vision-and-Language Navigation}
Vision-and-Language Navigation (VLN)~\cite{VLN} is a challenging task and has received continuous and intense attention from the academic community in recent years~\cite{Weakly-Supervised, Room-and-Object, Cross-Modal, Contrastive-PI, One-Step, CLEAR, DUET, Outdoor, ding2022embodied}. Early methods attempt to learn the agent from sequence-to-sequence models~\cite{VLN, Speaker-Follower, EnvDrop}. However, these methods can not model the cross-modal relation between language and visual observation well. To address this issue, transformer~\cite{Transformer} architecture is adopted to the agents followed by vision-and-language pre-training~\cite{SMNA, RCMS, AuxRN, VLNBert, Airbert, HOP, HAMT, SOAT, SEA}. PREVALENT~\cite{image-text-action} pre-trains transformer-based agent via masked language modeling and action prediction tasks. Inspired by BERT~\cite{BERT}, several works propose to use different variants of BERT for VLN pre-training. VLN-BERT~\cite{VLNBert} utilizes image-text data~\cite{ViLBERT} to perform path-instruction matching pretext task. Airbert~\cite{Airbert} proposes the shuffle loss to improve the ability of the model to learn the order of image-caption pairs. Recently, HOP~\cite{HOP} introduces the history and order-aware pretext tasks. However, these existing pretext tasks do not consider the learning of environment layout reasoning ability and bring limited performance for VLN tasks. In this work, we propose a trajectory judgment task to teach the agent to distinguish reasonable navigation trajectories. By proficiently accomplishing this task, the agent can acquire the ability to reason about environment layout, which enhances its generalization capability to unseen environments.

\subsection{Datasets for VLN}
The major difficulty of generalizing a VLN agent to unseen environments is the scarcity of VLN training data. Well-labeled VLN datasets built from the simulators~\cite{MP3d, HMP3D, Gibson} allow the agent to obtain a promising performance, such as R2R~\cite{VLN}, R4R~\cite{R4R}, RxR~\cite{RxR} and SOON~\cite{SOON}. While the data built from simulators are laborious, Wang \textit{et al.}~\cite{Less-is-More} and ProbES~\cite{Prompt} enrich the navigation instructions via self-exploration in the simulation environments. Some other endeavors~\cite{EnvDrop, Envedit, Speaker-Follower, APS, Pathdreamer, SIVLN} seek to augment the data from existing datasets. AutoVLN~\cite{AutoVLN} and Kamath \textit{et al.}~\cite{ANewPath} enlarges the VLN data from simulations with a larger number of environments. However, these datasets are limited by the number of scenes in simulators. To ease this problem, VLN-BERT~\cite{VLNBert} leverages the abundant web image-captions pairs as VLN pre-training data. Airbert~\cite{Airbert} further exploits indoor house images and captions from the web to construct path-instruction pairs for VLN pre-training. However, the trajectories constructed by simply splicing pictures may be confusing and ambiguous. In our work, we address these problems by proposing a large-scale in-domain VLN-like pre-training dataset, providing the agent with diverse visual environments and reasonable layouts.

\section{Building VLN Dataset from YouTube Videos}
\label{sec:dataset}

Our first step is to develop a large-scale VLN-like dataset that comprises reasonable path-instruction pairs from house tour videos on YouTube, termed YouTube-VLN. YouTube-VLN serves as a cornerstone for facilitating the acquisition of VLN capabilities for Lily, featuring diverse environments, real layouts and native actions. To achieve this goal, we present an entropy-based trajectory generation technique~(Section~\ref{sec:path_gen}) and an action-aware generator~(Section~\ref{sec:ins_gen}) to tackle the arduous tasks of trajectory and instruction generation, respectively.

\subsection{Diverse Trajectory Generation.}
\label{sec:path_gen}
We seek to construct discrete navigation trajectories from YouTube videos. Similar to discrete navigation datasets (\eg R2R~\cite{VLN}), each trajectory comprises $K$ navigation nodes, representing different locations of a navigation path. 
This entails addressing two major challenges:
1) how to determine the locations composing a trajectory to make the trajectory more diverse and
2) how to represent the visual content at each node location.
To tackle these challenges, we first collect large-scale consecutive indoor frames from YouTube videos. Then, we group the adjacent frames according to their room types and consider each group as a node. Last, we present an entropy-based technique to select the most informative frames in a group for representing the visual content in a node.

\begin{figure}[t]
	\centering
	\includegraphics[width=1\linewidth]{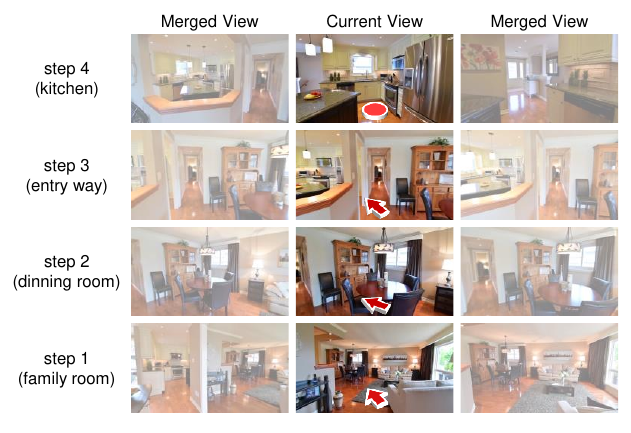}
	\caption{An example of a generated trajectory from a YouTube house tour video. \includegraphics[width=1.0em]{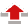} represents the direction to go next while~\includegraphics[width=1.0em]{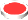} represents the stop action.}
	\label{fig:merge_strategy}
	\vspace{-4mm}
\end{figure}

\textbf{Collecting Navigation Data from YouTube.}
\label{sec:Collecting_In-door_Images}
Real estate agents typically tour a house in each video. To satisfy the visual diversity and dataset scale, we create the YouTube-VLN dataset from 4078 videos collected from various uploaders, with a total duration of 433 hours. In contrast to prior work~\cite{YTb}, which relied on a limited set of videos from a single uploader, our dataset features greater diversity and volume. We also employ sparse sampling and off-the-shelf image classifiers~\cite{Mask_RCNN, CLIP} to pre-process the videos, filtering out redundant or noisy frames (those featuring people or outdoor scenes), resulting in a final set of 587k indoor frames suitable for constructing trajectories.

\textbf{Determining the Locations of Trajectory Nodes.}
\label{sec:Grouping-Nodes}
A real robot often needs to go through different locations for navigating to a goal. To mimic the real navigation process, we expect that our constructed trajectories also contain diverse visual content within the limited navigation nodes. To achieve this, we first utilize the powerful large model CLIP~\cite{CLIP} to recognize the room type of each indoor image. Then, we gather temporally adjacent frames with the same room type as a group and consider this group as one of the navigation nodes. In this sense, the navigation nodes are diversely spread in different rooms and the constructed trajectories are able to mimic the real navigation process. We also call this kind of node a \textit{room node}. In practice, to increase the visual diversity of trajectories, we also randomly insert \textit{transition nodes} that are composed of video frames captured during the transition from one room to another one, between two adjacent room nodes.

\textbf{Representing Visual Content in a Node.}
\label{sec:landmarks}
A node consists of a group of images and sometimes the number of images may exceed 100 as the photographer could stay in the same room for a long time. Hence, we have to select the most informative images for representing node features. Inspired by EATA~\cite{EATA}, an image with lower classification entropy is more reliable, containing more information relevant to a specific class (room type in our case). We thus propose to select an image with the lowest classification entropy to represent the current view of a node. In order to mimic the panoramic visual context, we then merge $M$ adjacent consecutive images of the current view. It is worth noting that our node features better represent a panoramic view compared with Airbert~\cite{Airbert} because we merge adjacent frames that belong to the same place as the current view.

Ultimately, we randomly chose $K$ continuous nodes to construct a trajectory. An example of the constructed trajectory is shown in Figure~\ref{fig:merge_strategy}.

\subsection{Action-Aware Instruction Generation}
\label{sec:ins_gen}

In addition to constructing navigation trajectories, one more important step for building a VLN dataset is to create the corresponding instructions without manual annotation.
The main challenge for this step is how to correctly describe visual content and actions along navigation paths.
To conquer this challenge, we first generate instruction templates with verb and noun phrase blanks. Then, we describe each node in trajectories using the CLIP~\cite{CLIP} model and infer the native actions using an action inverse model~\cite{YTb}. To generate the final instructions, we fill the instruction templates with these visual descriptions and actions.

Specifically, we first generate templates with verb and noun blanks from instructions in the R2R dataset following Airbert~\cite{Airbert}.
For noun blanks, we fill them with visual content descriptions about each node. We select the frame with the lowest classification entropy (as described in Sec.~\ref{sec:landmarks}) and use the CLIP model to infer the objects it contains, together with the room type to populate a noun blank.
For verb blanks, the existing instruction generation method~\cite{Airbert} is unable to fill them with the correct action words because it cannot figure out the actions taken for navigating from one image to another. This makes the agent confused when it observes similar viewpoints transition but is given different action descriptions.
To tackle this problem, we propose an action-aware strategy to fill instruction templates with native actions instead of random inconsistent actions. To be specific, we follow~\cite{YTb} to train an action inverse model, which has 96\% prediction accuracy for predicting native actions, to pseudo-label the trajectory with action labels from one location node to another. The predicted actions are then converted into actionable verbs,~\ie ``go forward'', ``turn left'' and ``turn right''. For each noun blank that has been filled with the description of one node, we find its closest verb phrase blank and fill it with the pseudo-labeled action which is executed to reach the next node. This eventually enables us to create action-aware instructions.

 \begin{figure*}[t!]
	\centering
	\includegraphics[width=1\linewidth]{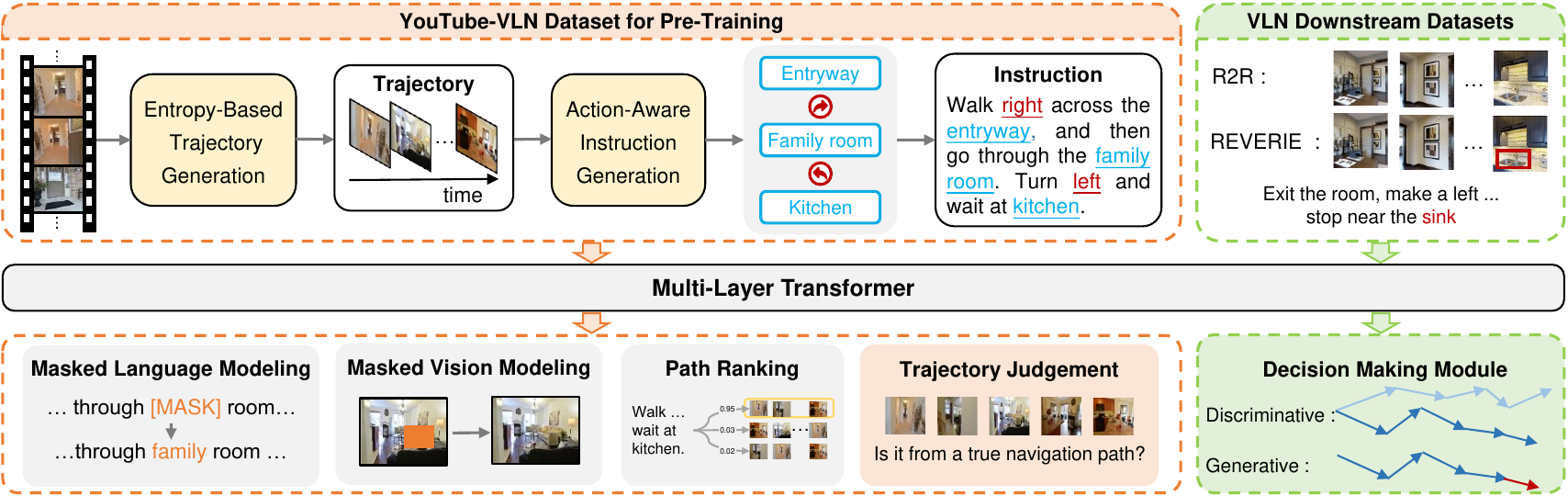}
	\caption{General scheme of our Lily agent for pre-training and downstream stages of VLN tasks. Lily agent learns to leverage the proposed YouTube-VLN dataset for the trajectory judgment task together with the other pretext tasks. After pre-training the VLN model, we adapt it to the downstream datasets and conduct different VLN downstream tasks.}
	\label{fig:scheme}
\end{figure*}

\section{Learning VLN from YouTube Videos}
\label{sec:approach}

Given the VLN-like and reasonable path-instruction pairs generated from YouTube videos, we then describe how to learn the Lily agent from these data. As shown in Figure~\ref{fig:scheme}, our VLN model consists of two components: a vision-and-language backbone (\ie Multi-Layer Transformer) that models the relationship between trajectories and instructions and a decision-making module that predicts the next action or a matching score for a  path-instruction pair. The vision-and-language backbone can be any type of cross-modal network. We chose ViLBERT~\cite{ViLBERT} for a fair comparison with Airbert~\cite{Airbert}. As a common practice, pretext tasks are utilized for pre-training the backbone. We next describe how to pre-train the backbone on our Youtube-VLN dataset using the proposed trajectory judgment pretext task. 

\subsection{Model Architecture}
\label{sec:Architecture}
We follow Airbert~\cite{Airbert} to leverage a ViLBERT-like~\cite{ViLBERT} architecture as the model backbone. The model encodes the sequential visual region features and the text token via two separate transformers respectively. More formally, the  path-instruction pair consists of $K$ nodes $\left\{V_k\right\}_{k=1}^K$ and $L$ text tokens $\left\{w_l\right\}_{l=1}^L$. Each node $V_k$ is composed of $R_k$ visual region features $\left\{v_i^k\right\}_{i=1}^{R_k}$. In this way, we represent the visual and text inputs respectively as follows:
\begin{small}
\begin{equation}
X_V=\left[[\texttt{IMG}], v_1^1, \ldots, v_{R_1}^1, \ldots,[\texttt{IMG}], v_1^K, \ldots, v_{R_K}^K\right],
\label{eq:xv}
\end{equation}
\end{small}
\vspace{-4mm}
\begin{small}
\begin{equation}
X_W=\left[[\texttt{CLS}], w_1, \ldots, w_l, \ldots, w_L,[\texttt {SEP}]\right],
\label{eq:xw}
\end{equation}
\end{small}

\noindent where [\texttt{IMG}], [\texttt{CLS}] and [\texttt{SEP}] are special tokens.
The encoded visual and text tokens finally interact via a cross-modal transformer encoder. We represent the whole model architecture as ``Multi-Layer Transformer'' in Figure~\ref{fig:scheme}.
 
\subsection{Learning Layout from Trajectory Judgment}
\label{sec:Pre-training}

Given the aforementioned model architecture, we propose to train a VLN agent with a trajectory judgment (TJ) task, enabling it to reason about layouts. Herein, we elaborate on the proposed trajectory judgment task.

\textbf{Formulation.} 
The trajectory judgment task aims to judge the reasonableness of trajectories. We consider the trajectories generated in the way described in Section~\ref{sec:path_gen} as positive (reasonable) samples and the shuffled trajectories as negative (unreasonable) ones. To finish this task, the agent is required to reason about the visual information and identify the room types, then infer whether the trajectory matches the real environment layout distribution. 
Specifically, we first calculate the dot product of the output features of [\texttt{IMG}] and [\texttt{CLS}] tokens. Then, we feed this vector feature to a linear layer to predict the probability that indicates whether the trajectory is reasonable.
The model aims to minimize the binary cross-entropy loss:

\vspace{-4mm}
\begin{small}
\begin{equation}
L=-\frac{1}{N} \sum_{n=1}^N\left[w \cdot y_n \log \left(p_n\right) + (1-y_n) \log \left({1-p_n}\right)\right],
\label{eq:tj_loss}
\end{equation}
\end{small}
\vspace{-4mm}

\noindent where $y_n = 1$ if the $n^{th}$ trajectory is reasonable, otherwise $y_n = 0$. $p_n$ represents the probability that the $n^{th}$ trajectory is predicted as reasonable. $N$ is the number of trajectories in a batch.
$w$ is a factor to mitigate the imbalance of positive and negative samples, which equals the ratio of the number of negative samples to the number of positive samples.

\textbf{Sample Generation.}
 We propose to shuffle the positive sample to generate the negative samples: 1) shuffle only the transition nodes; 2) shuffle all the nodes; 3) keep the order of the room nodes, and randomly insert nodes from other videos. In this way, we create rich and hard negative samples, which increases the task difficulty, helping the agent understand the real layout in a more complex manner.

\textbf{Combining with Existing Pre-Training Tasks.}
As depicted in Figure~\ref{fig:scheme}, we follow Airbert~\cite{Airbert} to pre-train the model backbone using our proposed trajectory judgment task, additionally combining with three other existing pretext tasks, namely masked language modeling (MLM), masked vision modeling (MVM) and path ranking (PR) on YouTube-VLN dataset. For MLM, we randomly mask out the words in instruction and the goal is to recover the masked words. Similar to MLM, MVM is designed to predict masked image regions. PR is a ranking task, which aims to decide the most matching path-instruction pair among a few pairs.

\begin{table*}[t!]
{
\centering
\resizebox{1.0\linewidth}{!}{
\begin{tabular}{clccclclccccclccccc}
\topline
\multirow{2}{*}{\#} &  & \multicolumn{3}{c}{Dataset}  &  & Pre-training Task &  & \multicolumn{5}{c}{Val Seen} &  & \multicolumn{5}{c}{Val Unseen}\\ \cmidrule{3-5} \cmidrule{7-7} \cmidrule{9-13} \cmidrule{15-19} 
 &  & Source  & \begin{tabular}[c]{@{}c@{}}Reasonable \\Navigation Path\end{tabular} & \begin{tabular}[c]{@{}c@{}}Pseudo-labeled \\ Action\end{tabular} &  & \begin{tabular}[c]{@{}c@{}}Trajectory\\ Judgment\end{tabular} 
                                                              &   & TL           & NE↓          & OSR↑          & SR↑           & SPL↑          &  & TL          & NE↓          & OSR↑          & SR↑           & SPL↑          \\ 
\midline
1    &  & Airbnb Images    &\XSolidBrush &\XSolidBrush &  &\XSolidBrush &   & 10.21 & 3.41 & 79.02 & 74.12 & 0.70 &  & 9.63 & 3.95 & 70.97 & 62.84 & 0.58 \\ 
\midline
2    &  & YouTube Videos &\XSolidBrush &\XSolidBrush &  &\XSolidBrush &   & 10.12 & 3.40 & 79.90 & 74.31 & 0.70 &  & 9.81 & 3.72 & 74.24 & 63.73 & 0.59 \\
3    &  & YouTube Videos & \Checkmark  &\XSolidBrush &  &\XSolidBrush &   & 10.30 & 3.40 & 78.60 & 75.10 & 0.71 &  & 9.60 & 3.70 & 73.50 & 65.00 & 0.61 \\
4    &  & YouTube Videos & \Checkmark  & \Checkmark  &  &\XSolidBrush &   & 10.20 & 3.30 & 79.80 & 75.40 & 0.71 &  & 9.30 & 3.60 & 72.70 & 66.10 & 0.62 \\
5    &  & YouTube Videos & \Checkmark  & \Checkmark &  & \Checkmark   &   & 9.99  & \textbf{3.12} & \textbf{80.88} & \textbf{77.45} & \textbf{0.74} &  & 9.64 & \textbf{3.37} & \textbf{74.93} & \textbf{66.70} & \textbf{0.62} \\ 
\bottomline
\end{tabular}
}
}
\caption{Ablation study on YouTube-VLN dataset and trajectory judgment pretext task for pre-training.}
\label{tab:ablation}
\vspace{-4mm}
\end{table*}

\subsection{Adapting Pre-trained Backbone for VLN}
\label{sec:Downstream}
We adapt the pre-trained model to both goal-oriented navigation task and object-oriented navigation task. All the tasks are based on the Matterport3D simulator~\cite{MP3d}. We utilize R2R~\cite{VLN} as the benchmark for the goal-oriented navigation task, which is divided into discriminative setting and generative setting. As for the object-oriented task, we evaluate our model on REVERIE~\cite{REVERIE} in generative setting. 

The discriminative setting formulates VLN as a path-selection problem, requiring the agent to choose the path that best matches the instruction from multiple candidate paths. Under the discriminative setting, we utilize the classifier used in the path ranking pretext task for decision-making and fine-tune the Lily agent on the R2R dataset. 

In the generative setting, the agent needs to predict actions sequentially to reach the goal (R2R) or simultaneously find the object (REVERIE). We adopt DUET~\cite{DUET} as the architecture for fine-tuning, which feeds the cross-modal feature into a feed-forward network for decision-making. We initialize the text transformer encoder and cross-modal transformer encoder of the generative model with our Lily agent. 
Note that our Lily agent can apply to any generative model. More details are available in the supplementary.

\section{Experiments}
\subsection{Experimental Setup}

\textbf{Dataset and Evaluation Metrics.}
We conduct our experiments on two VLN benchmarks,~\ie R2R~\cite{VLN} and REVERIE~\cite{REVERIE}. These two datasets consist of 21,567 path-instruction pairs from 90 scenes in Matterport3D~\cite{MP3d}. REVERIE follows the same train/val/test splits as the R2R, while requiring an agent to select the bounding box 
 of the target object bounding box additionally. Following standard settings~\cite{Airbert}, we adopt five metrics for evaluating R2R, namely success rate \textbf{(SR)}, oracle success rate \textbf{(OSR)}, success rate weighted by the ratio between the length of the shortest path and the predicted path \textbf{(SPL)}, trajectory length \textbf{(TL)} and navigation error \textbf{(NE)}. As for REVERIE, we leverage four metrics for evaluating navigation performance, namely \textbf{SR}, \textbf{OSR}, \textbf{SPL} and \textbf{TL}, and two for object grounding performance, namely remote grounding success (\textbf{RGS}) and RGS weighted by path length (\textbf{RGSPL}).

\textbf{Implementation Details.} 
We implement our method based on Pytorch framework~\cite{pytorch} and Matterport3D simulator~\cite{MP3d}. Specifically, we divide our training process into two stages,~\ie pre-training and fine-tuning. For the pre-training stage, we distribute training over 4 NVIDIA 3090 GPUs for 500 epochs to convergence.  
The pre-trained model with the highest accuracy for the path ranking pretext task is selected for fine-tuning.
During the fine-tuning stage, we distribute training over 8 NVIDIA 3090 GPUs for 30 epochs to convergence. 
Following Airbert~\cite{Airbert}, we use augmented data from EnvDrop~\cite{EnvDrop} for fine-tuning by default.
More details are provided in the supplementary.

\subsection{Ablation Studies on Pre-Training}
\label{sec:ablstion}
We ablate our approach under discriminative setting on R2R benchmark. Considering the time efficiency, we do not use augmented data for fine-tuning on these experiments.

\textbf{Data Source: Airbnb Images~\vs YouTube Videos.}
One of the main differences between YouTube-VLN dataset and Airbnb dataset~\cite{Airbert} is the data source. Airbnb consists of 713k image-caption pairs while YouTube-VLN consists of 587k images extracted from 433 hours of house tour videos.
YouTube-VLN has fewer images but provides more information about a room from different camera angles by merging, which better simulates a panorama for downstream VLN tasks. 
To evaluate the effect of data source, we follow Airbert~\cite{Airbert} to randomly select images in the same house
to build a trajectory and its corresponding instruction.
We keep the number of instructions the same in pre-training for a fair comparison.
In Table~\ref{tab:ablation} (\# 1 \vs \# 2), YouTube data performs slightly better than Airbnb data, surpassing the SR by 0.89\% on the val unseen split.  We speculate this is because data quality is more important than quantity.

\textbf{Effectiveness of Reasonable Navigation Trajectory.}
YouTube-VLN dataset is collected from real house tour videos and is thus able to extract frames in chronological order to build reasonable navigation trajectories instead of combining multiple randomly chosen images.
We use the generated reasonable navigation trajectories and shuffled trajectories to train two agents, respectively. In Table~\ref{tab:ablation}, the agent trained with reasonable navigation trajectories (\# 3) achieves significantly better performance than the shuffled navigation trajectories variant (\# 2), with 1.27\% gains on SR under the val unseen split. This suggests that the agent can not well understand and ground the instruction to the trajectory without reasonable navigation paths for learning.

\textbf{Effectiveness of Pseudo-Labeled Action.}
The core of the proposed action-aware instruction generator is the pseudo-labeled actions for the instructions. To evaluate the effectiveness of the pseudo-labeled actions, we construct a variant that replaces random action words with the pseudo-labeled actions which are filled in the instructions. The results are shown in Table~\ref{tab:ablation}, \# 4. Compared to the variant (\# 3) that fills the instructions with random action words, this variant boosts the SR metric on both the val unseen (+1.10$\%$) and seen (+0.30$\%$) splits, showing the effectiveness of pseudo-labeled actions. It indicates that with pseudo-labeled actions, the agent can effectively ground the action to the visual observation and recognizes the correct transition from one location to another.

\textbf{Effectiveness of Trajectory Judgment Pretext Task.}
To explore the layout knowledge of the YouTube-VLN dataset and equip the agent with layout reasoning ability, we propose a self-supervised trajectory judgment pretext task. In Table~\ref{tab:ablation}, the variant with the proposed pretext task improves the performance (\# 5 \vs \# 4, +2.05\% on val seen and +0.60\% on val unseen \textit{w.r.t.} SR). Moreover, given a room type and visual information of the current node as inputs, this variant predicts the relative orientation of nodes for that room type with a 30\% increase in accuracy (see supplementary for more details). These substantiate the claim of the importance of the proposed trajectory judgment task, which helps the agent learn the layout reasoning ability.

\begin{table}[!t]
{
\centering
\resizebox{1.0 \linewidth}{!}{
\begin{tabular}{clllccccc}
\topline
        \#          &  &  Methods                 &  & TL            & NE↓           & OSR↑          & SR↑             & SPL↑      \\ \midline
1                   &  & Random Sample           &  & 9.37 & 3.52 & 72.88 & 64.41 & 0.61 \\
2                   &  & Temporal Difference     &  & 9.55 & 3.61 & 73.35 & 65.30 & 0.61 \\
3                   &  & Entropy-Based           &  & 9.64 & \textbf{3.37} & \textbf{74.93} & \textbf{66.70} & \textbf{0.62}\\ \bottomline
\end{tabular}
}
}
\caption{Comparison between different strategies of selecting frames to represent nodes under val unseen split.}
\label{tab:KeyFrames}
\vspace{-4mm}
\end{table}

\textbf{Effectiveness of Entropy-Based Trajectory Generation Strategy.}
In order to acquire an informative frame to represent a node, we propose an entropy-based technique as mentioned in Section~\ref{sec:path_gen}. To demonstrate the effectiveness of our strategy, we construct two variants,~\ie one randomly chooses a frame, and one decides the frame by temporal difference~\cite{TD}. The second variant computes the temporal pixel difference between every two consecutive frames and picks the frames where the peaks are located as the frames to represent the nodes. In Table~\ref{tab:KeyFrames}, our entropy-based method significantly outperforms these two variants, increasing the SR on the val unseen split from 64.41\% and 65.30\% to 66.70\%, respectively. We speculate this is because 1) randomly sampling frames can generate consecutive redundant frames or meaningless frames (\eg a wall takes up most of the frame); 2) the temporal difference method selects a frame that 
represents the junction of two different rooms, which is ambiguous and thus confuses the agent. In comparison, our entropy-based method is able to find a reliable frame to represent a node and ensure two adjacent nodes belong to different room types.

\subsection{Comparison with State-of-the-Arts}

\textbf{Results on R2R Dataset.}
We first compare our Lily agent with current methods under the discriminative setting. In Table~\ref{tab:discriminative}, compared with VLN-BERT that uses image-caption pairs from the web for pre-training, our Lily agent significantly increases the SR by 10.74\% on the val unseen split. This highlights the importance of providing the in-domain indoor data for pre-training. Moreover, compared with Airbert which uses in-domain image-caption pairs from online rental marketplaces, Lily still increases the SR from 73.85\% to 79.31\% on val seen and from 68.67\% to 70.00\% on val unseen. We attribute the improvement to our proposed VLN-like YouTube-VLN dataset and trajectory judgment pretext task, which have been thoughtfully evaluated in Section~\ref{sec:ablstion}. When ensembled with the speaker-follower~\cite{Speaker-Follower}, all three methods increase the performance and our Lily agent performs the best.
\begin{table}[!t]
{
\resizebox{1.0\linewidth}{!}{

    \begin{tabular}{lllcccccccccccc}
    \topline
    \multicolumn{1}{l}{\multirow{2}{*}{Methods}} & \multicolumn{4}{c}{Val Seen}    & \multicolumn{4}{c}{Val Unseen}                                          \\ \cmidrule{2-9}
                       \multicolumn{1}{c}{}     & TL            & NE↓           & SR↑            & SPL↑           & TL         & NE↓           & SR↑            & SPL↑           \\ \midline
    Follower~\cite{Speaker-Follower}          & 10.40          & 3.68                    & 65.10          & 0.62          & 9.57       & 5.20                    & 52.36          & 0.49          \\
    Speaker~\cite{Speaker-Follower}           & 11.19          & 3.80                    & 60.69          & 0.56          & 10.71      & 4.25                    & 54.66          & 0.49          \\
    \footnotesize{Speaker-Follower}~\cite{Speaker-Follower}  & 10.69          & 2.72                    & 74.22          & 0.70          & 10.10      & 3.32                    & 67.90          & 0.63          \\
    ProbES~\cite{Prompt}                      & -              & -                       & -              & -             & 9.50       & 4.05                    & 60.28          & 0.56          \\
    VLN-BERT~\cite{Improving}                 & 10.28          & 3.73                    & 70.20          & 0.66          & 9.60       & 4.10                    & 59.26          & 0.55          \\
    Airbert~\cite{Airbert}                    & 10.59          & 3.21                    & 73.85          & 0.69          & 10.03      & 3.24                    & 68.67          & 0.63          \\
    Lily                                      & 10.21          & 2.89                    & \textbf{79.31}  & \textbf{0.76}     & 10.03      & \textbf{3.19}             & \textbf{70.00}  & \textbf{0.65}     \\ \midline
    VLN-BERT*~\cite{Improving}                & 10.61          & 2.35                    & 81.86          & 0.78          & 10.00      & 2.76                   & 73.61          & 0.68          \\
    Airbert*~\cite{Airbert}~                  & 10.63          & 2.13                    & 81.40          & 0.77          & 9.99       & 2.69                    & 75.01          & 0.70          \\
    Lily*                                    & 10.51           & \textbf{2.06}           & \textbf{83.29}          & \textbf{0.80}   & 9.78  & \textbf{2.48}             & \textbf{76.88}  & \textbf{0.72} \\ 
    \bottomline
    \end{tabular}
}
}
\centering
\caption{Comparison with state-of-the-arts on R2R dataset under discriminative setting. * means results of ensembling with the speaker-follower~\cite{Speaker-Follower} model.}
\label{tab:discriminative}
\vspace{-2mm}
\end{table}

\begin{table}[t]
\centering
\resizebox{1.0\linewidth}{!}{
\begin{tabular}{lccccc}
\topline
           Methods              & TL     & NE↓  & SPL↑ & OSR↑ & SR↑  \\ \midline

Speaker-Follower~\cite{Speaker-Follower}         & 1257    & 4.87 & 0.01  & 96   & 53   \\
Self-Monitoring ~\cite{SMNA}                     & 373     & 4.48 & 0.02  & 97   & 61   \\
Reinforced CM   ~\cite{RCM}                      & 358     & 4.03 & 0.02  & 96   & 63   \\
EnvDrop         ~\cite{EnvDrop}                  & 687     & 3.26 & 0.01  & 99   & 69   \\
AuxRN           ~\cite{AuxRN}                    & 41      & 3.24 & 0.21  & 81   & 71   \\
VLN-BERT~\cite{Improving}                        & 686.82  & 2.99 & 0.01  & 99   & 73   \\
Global Normalization~\cite{GN}                   & 686.86  & 2.99 & 0.01  & 99   & 74   \\
Airbert~\cite{Airbert}                           & 686.54  & 2.58 & 0.01  & 99   & 77 \\ \midline
LiLy                                             & 686.45 & \textbf{2.50}       & 0.01   &  99    & \textbf{79}     \\ \bottomline
\end{tabular}
}
\setlength{\abovecaptionskip}{0.15cm}
\caption{Results under discriminative setting on the test unseen split as indicated on the R2R leaderboard~\protect\footnotemark.}
\label{tab:leaderboard}
\vspace{-4mm}
\end{table}

\footnotetext{\fontsize{5.75pt}{1pt}{\url{https://eval.ai/web/challenges/challenge-page/97/leaderboard/270}}}

\begin{table*}[t!]
\centering
\resizebox{0.8\linewidth}{!}{\begin{tabular}{lp{0.8cm}<{\centering}p{0.8cm}<{\centering}p{0.8cm}<{\centering}p{0.8cm}<{\centering}cp{0.8cm}<{\centering}p{1.0cm}<{\centering}cp{0.8cm}<{\centering}p{0.8cm}<{\centering}p{0.8cm}<{\centering}p{0.8cm}<{\centering}cp{0.8cm}<{\centering}p{1.0cm}<{\centering}cp{0.8cm}<{\centering}p{0.8cm}<{\centering}p{0.8cm}<{\centering}p{0.8cm}<{\centering}cp{0.8cm}<{\centering}p{1.0cm}<{\centering}}
\topline
\multirow{3}{*}{Methods} & \multicolumn{7}{c}{Val Unseen}                                                     &  & \multicolumn{7}{c}{Test Unseen}             \\ 
                           \cmidrule{2-8} \cmidrule{10-16}
                         & \multicolumn{4}{c}{Navigation}                  &  & \multicolumn{2}{c}{Grounding} &  & \multicolumn{4}{c}{Navigation} &  & \multicolumn{2}{c}{Grounding} \\ \cmidrule{2-5} \cmidrule{7-8} \cmidrule{10-13} \cmidrule{15-16}
                         & TL   & OSR↑  & SR↑            & SPL↑           &  & RGS↑          & RGSPL↑        &  & TL    & OSR↑  & SR↑   & SPL↑  &  & RGS↑          & RGSPL↑         \\ \midline
Human                    & -     & -     & -              & -              &  & -             & -            &  & 21.18  & 86.83 & 81.53 & 83.66 &  & 77.84         & 51.44          \\ \midline
Seq2Seq~\cite{VLN}       & 11.07 & 8.07  & 4.20           & 2.84           &  & 2.16          & 1.63          &  & 10.89  & 6.88  & 3.99  & 3.09  &  & 2.00          & 1.58           \\
RCM~\cite{RCM}           & 11.98 & 14.23 & 9.29           & 6.97           &  & 4.89          & 3.89          &  & 10.60  & 11.68 & 7.84  & 6.67  &  & 3.67          & 3.14           \\
SMNA~\cite{SMNA}         & 9.07  & 11.28 & 8.15           & 6.44           &  & 4.54          & 3.61          &  & 9.23   & 8.39  & 5.80  & 4.53  &  & 3.10          & 2.39           \\
FM~\cite{REVERIE}        & 45.28 & 28.20 & 14.40          & 7.19           &  & 7.84          & 4.67          &  & 39.05  & 30.63 & 19.88 & 11.61 &  & 11.28         & 6.08           \\
SIA~\cite{SIA}           & 41.53 & 44.67 & 31.53          & 16.28          &  & 22.41         & 11.56         &  & 48.61  & 44.56 & 30.80 & 14.85 &  & 19.02         & 9.20           \\
HAMT~\cite{HAMT}         & 14.08 & 36.84 & 32.95          & 30.20          &  & 18.92         & 17.28         &  & 13.62  & 33.41 & 30.40 & 26.67 &  & 14.88         & 13.08           \\
RecBERT~\cite{RecVLN}    & 16.78 & 35.02 & 30.67          & 24.90          &  & 18.77         & 15.27         &  & 15.86  & 32.91 & 29.61 & 23.99 &  & 16.50         & 13.51          \\
ProbES~\cite{Prompt}     & 18.00 & 33.23 & 27.63          & 22.75          &  & 16.84         & 13.94         &  & 16.84  & 28.23 & 24.97 & 20.12 &  & 15.11         & 12.32        \\ 
Airbert~\cite{Airbert}   & 18.71 & 34.51 & 27.89          & 21.88          &  & 18.23         & 14.18         &  & 17.91  & 34.20 & 30.28 & 23.61 &  & 16.83        & 13.28          \\\cmidrule{1-16}
DUET~\cite{DUET}         & 22.11 & 51.07 & 46.98          & 33.73          &  & 32.15         & 23.03         &  & 21.30  & 56.91 & 52.51 & 36.06 &  & 31.88         & 22.06         \\
DUET~(Lily)              & 21.87 & \textbf{53.71} & \textbf{48.11} & \textbf{34.43} &  &\textbf{32.15} & \textbf{23.43}         &  & 21.94 & \textbf{60.51} & \textbf{54.32} & \textbf{37.34} &  & \textbf{32.02} & 21.94           \\ \bottomline
\end{tabular}
}
\setlength{\abovecaptionskip}{0.15cm}
\caption{Comparison with state-of-the-arts  on REVERIE. Lily agent achieves the state-of-the-art performance on all splits.}
\label{tab:REVERIE_SOTA}
\vspace{-4mm}
\end{table*}

\begin{table}[htb]
\centering
\resizebox{1.0\linewidth}{!}{\begin{tabular}{lcccccccc}
\topline
\multirow{2}{*}{Methods}                     & \multicolumn{4}{c}{Val Unseen}              &\multicolumn{4}{c}{Test Unseen}          \\ \cmidrule{2-6} \cmidrule{7-9}
                            & TL     & NE↓         & SR↑      & SPL↑      &TL     & NE↓        & SR↑     & SPL↑     \\ \cmidrule{1-9}
Seq2Seq~\cite{VLN}          & 8.39   & 7.81        & 22       & -         &8.13   & 7.85       & 20      & -        \\
EnvDrop~\cite{EnvDrop}      & 10.70  & 5.22        & 52       & 48        &11.66  & 5.23       & 51      & 47       \\
AuxRN~\cite{AuxRN}          & -      & 5.28        & 55       & 50        &-      & 5.15       & 55      & 51       \\
\footnotesize{PREVALENT}~\cite{Generic}    & 10.19  & 4.71        & 58       & 53        &10.51  & 5.30       & 54      & 51       \\
RelGraph~\cite{RelGraph}    & 9.99   & 4.73        & 57       & 53        &10.29  & 4.75       & 55      & 52       \\
RecBERT~\cite{RecVLN}       & 12.01  & 3.93        & 63       & 57        &12.35  & 4.09       & 63      & 57       \\
ProbES~\cite{Prompt}        & 11.58  & 4.00        & 61       & 55        &12.43  & 4.20       & 62      & 56       \\
ADAPT~\cite{ADAPT}          & 12.33  & 3.66        & 66       & 59        &13.16  & 4.11       & 63      & 57       \\
HOP~\cite{HOP}              & 12.27  & 3.80        & 64       & 57        &12.65  & 3.83       & 64      & 59       \\
HAMT~\cite{HAMT}            & 11.46  & 2.29        & 66       & 61        &12.27  & 3.93       & 65      & 60       \\
Airbert~\cite{Airbert}      & 11.78  & 4.01        & 62       & 56        &12.41  & 4.13       & 62      & 57       \\ \cmidrule{1-9}
DUET~\cite{DUET}            & 13.94  & 3.31        & 72       & 60        &14.73  & 3.65       & 69      & 59       \\ 
DUET (Lily)                 & 14.58  & \textbf{2.90}  & \textbf{74}  & \textbf{62} &16.13  & \textbf{3.44}  &\textbf{72}  & \textbf{60}  \\ \bottomline
\end{tabular}
}
\caption{Comparison with state-of-the-arts on R2R dataset under generative setting.}
\label{tab:generative}
\vspace{-3mm}
\end{table}

In Table~\ref{tab:leaderboard}, we evaluate on R2R test split and our Lily ranks first on the VLN challenge leaderboard compared with the results whose manuscripts are publicly available, achieving the highest SR of 79\%. As we follow Airbert to use 30 candidate trajectories from EnvDrop~\cite{EnvDrop} and the leaderboard considers that our agent has walked through all these paths, the SPL metric is low for both Lily and Airbert.

Besides, our Lily agent also helps to increase the performance on R2R under the generative setting. We enhance the state-of-the-art DUET~\cite{DUET} method by pre-training the agent using our method as mentioned in Section~\ref{sec:approach}. In Table~\ref{tab:generative}, the agent incorporating Lily achieves 2\% and 3\% improvements \textit{w.r.t.} SR on the val unseen split and test unseen split, respectively, compared to DUET. Notably, our method increases the SPL by 2\% on the val unseen split, indicating the agent is able to reach the goal more efficiently. We attribute this to the agent's acquisition of layout prior knowledge via the proposed trajectory judgment task on our YouTube-VLN dataset, enabling it to plan more efficient routes to the goals in new environments.

\textbf{Results on REVERIE Dataset.}
Compared with R2R, REVERIE is more challenging as its instructions only describe the destinations without detailed path descriptions. This requires the agent to be equipped with common knowledge about the room layouts and to reason the possible paths that lead to the destinations. In Table~\ref{tab:REVERIE_SOTA}, we outperform the SOTA agent (\ie DUET) by 1.13\% on SR and increase the SPL from 46.98\% to 48.11\% on the val unseen split. It is worth noting that our method achieves a higher OSR with a shorter TL, indicating that our agent finds the destination more quickly. We attribute this to the better layout reason ability learned from the large-scale diverse reasonable trajectories in the proposed YouTube-VLN dataset. A similar performance is obtained on the test unseen split, where the Lily agent improves the SR by 1.81\% and the SPL by 1.28\%. Although our objective is solely navigation, we still achieve comparable performance on grounding metrics and even slightly outstrip DUET on most metrics.

\subsection{Learning Navigation from One Environment}

Our intuition is that pre-training on the YouTube-VLN dataset can mitigate the domain gap of training from scratch and is able to achieve excellent performance with only a few training environments. To verify this, we conduct a one-shot learning study, where we fine-tune our model on only one environment of the original training environments. Note that the candidate paths are generated from all of the possible paths from the start viewpoints to all navigable points, instead of from the expert model in EnvDrop~\cite{EnvDrop}. All the candidate paths are the shortest paths in the navigation graphs. To reduce the bias, we randomly select 5 sets from the entire environments and report the average results.

In Table~\ref{tab:few-shot}, our agent outperforms all the existing pre-training methods. On the val seen split using one-shot fine-tuning, compared to VLN-BERT and AirBERT, our Lily agent achieves 3.60\% and 1.43\% improvements, respectively. In the val unseen split using one-shot fine-tuning, we achieve 28.43\% improvement compared to VLN-BERT. All these results suggest the effectiveness of our method.

 \begin{table}[t!]
\centering
\resizebox{0.6\linewidth}{!}{
\begin{tabular}{lcccccc}
\topline
Methods                   &         & Val Seen      & Val Unseen                   \\ \midline
VLN BERT~\cite{Improving} &         & 45.71         & 22.43                      \\
AirBERT~\cite{Airbert}    &         & 47.88         & 50.00                      \\
Lily                      &         & \textbf{49.31}& \textbf{50.86}             \\ \bottomline
\end{tabular}
}
\caption{SR on val seen and val unseen splits of R2R. All the agents access only one environment.}
\label{tab:few-shot}
\vspace{-4mm}
\end{table}

\section{Conclusion}

In this work, we propose a new approach Lily to address the limitations of existing vision-and-language navigation (VLN) methods by creating a large-scale VLN-like dataset from real house tour videos to train our embodied agent. We overcome the challenges of automatically generating path-instruction pairs to construct the dataset from raw and unlabeled videos by leveraging an entropy-based method for trajectory construction and an action-aware generator for instruction generation. Additionally, we train the agent to judge the reasonableness of trajectories, improving its layout reasoning ability. Our method achieves state-of-the-art performance on two popular benchmarks (R2R and REVERIE), demonstrating the efficacy. Overall, we hope our work can provide valuable insight into the VLN community by learning embodied VLN from passive videos.

\section*{Acknowledgement}
Prof. Tan and his students were partially supported by the 
National Natural Science Foundation of China (NSFC) (62072190), National Natural Science Foundation of China (NSFC) 61836003 (key project), Program for Guangdong Introducing Innovative and Enterpreneurial Teams 2017ZT07X183.

\bibliographystyle{abbrv}
{
	\small
	\bibliography{ref}

\begin{thebibliography}{10}

\bibitem{bottom-up}
P.~Anderson, X.~He, C.~Buehler, D.~Teney, M.~Johnson, S.~Gould, and L.~Zhang.
\newblock Bottom-up and top-down attention for image captioning and visual
  question answering.
\newblock In {\em {CVPR}}, pages 6077--6086, 2018.

\bibitem{VLN}
P.~Anderson, Q.~Wu, D.~Teney, J.~Bruce, M.~Johnson, N.~S{\"{u}}nderhauf, I.~D.
  Reid, S.~Gould, and A.~van~den Hengel.
\newblock Vision-and-language navigation: Interpreting visually-grounded
  navigation instructions in real environments.
\newblock In {\em {CVPR}}, pages 3674--3683, 2018.

\bibitem{MP3d}
A.~X. Chang, A.~Dai, T.~A. Funkhouser, M.~Halber, M.~Nie{\ss}ner, M.~Savva,
  S.~Song, A.~Zeng, and Y.~Zhang.
\newblock Matterport3d: Learning from {RGB-D} data in indoor environments.
\newblock In {\em {3DV}}, pages 667--676, 2017.

\bibitem{YTb}
M.~Chang, A.~Gupta, and S.~Gupta.
\newblock Semantic visual navigation by watching youtube videos.
\newblock In {\em {NeurIPS}}, pages 4283--4294, 2020.

\bibitem{Weakly-Supervised}
P.~Chen, D.~Ji, K.~Lin, R.~Zeng, T.~H. Li, M.~Tan, and C.~Gan.
\newblock Weakly-supervised multi-granularity map learning for
  vision-and-language navigation.
\newblock In {\em {NeurIPS}}, 2022.

\bibitem{HAMT}
S.~Chen, P.~Guhur, C.~Schmid, and I.~Laptev.
\newblock History aware multimodal transformer for vision-and-language
  navigation.
\newblock In {\em {NeurIPS}}, pages 5834--5847, 2021.

\bibitem{DUET}
S.~Chen, P.~Guhur, M.~Tapaswi, C.~Schmid, and I.~Laptev.
\newblock Think global, act local: Dual-scale graph transformer for
  vision-and-language navigation.
\newblock In {\em {CVPR}}, pages 16516--16526, 2022.

\bibitem{AutoVLN}
S.~Chen, P.-L. Guhur, M.~Tapaswi, C.~Schmid, and I.~Laptev.
\newblock Learning from unlabeled 3d environments for vision-and-language
  navigation.
\newblock In {\em {ECCV}}, 2022.

\bibitem{BERT}
J.~Devlin, M.~Chang, K.~Lee, and K.~Toutanova.
\newblock {BERT:} pre-training of deep bidirectional transformers for language
  understanding.
\newblock In {\em {NAACL}}, pages 4171--4186, 2019.

\bibitem{ding2022embodied}
M.~Ding, Y.~Xu, Z.~Chen, D.~D. Cox, P.~Luo, J.~B. Tenenbaum, and C.~Gan.
\newblock Embodied concept learner: Self-supervised learning of concepts and
  mapping through instruction following.
\newblock In {\em CoRL}, 2022.

\bibitem{Speaker-Follower}
D.~Fried, R.~Hu, V.~Cirik, A.~Rohrbach, J.~Andreas, L.~Morency,
  T.~Berg{-}Kirkpatrick, K.~Saenko, D.~Klein, and T.~Darrell.
\newblock Speaker-follower models for vision-and-language navigation.
\newblock In {\em {NeurIPS}}, pages 3318--3329, 2018.

\bibitem{APS}
T.~Fu, X.~E. Wang, M.~F. Peterson, S.~T. Grafton, M.~P. Eckstein, and W.~Y.
  Wang.
\newblock Counterfactual vision-and-language navigation via adversarial path
  sampler.
\newblock In {\em {ECCV}}, pages 71--86, 2020.

\bibitem{Room-and-Object}
C.~Gao, J.~Chen, S.~Liu, L.~Wang, Q.~Zhang, and Q.~Wu.
\newblock Room-and-object aware knowledge reasoning for remote embodied
  referring expression.
\newblock In {\em {CVPR}}, pages 3064--3073, 2021.

\bibitem{Airbert}
P.~Guhur, M.~Tapaswi, S.~Chen, I.~Laptev, and C.~Schmid.
\newblock Airbert: In-domain pretraining for vision-and-language navigation.
\newblock In {\em {ICCV}}, pages 1614--1623, 2021.

\bibitem{Generic}
W.~Hao, C.~Li, X.~Li, L.~Carin, and J.~Gao.
\newblock Towards learning a generic agent for vision-and-language navigation
  via pre-training.
\newblock In {\em {CVPR}}, pages 13134--13143, 2020.

\bibitem{image-text-action}
W.~Hao, C.~Li, X.~Li, L.~Carin, and J.~Gao.
\newblock Towards learning a generic agent for vision-and-language navigation
  via pre-training.
\newblock In {\em {CVPR}}, pages 13134--13143, 2020.

\bibitem{Mask_RCNN}
K.~He, G.~Gkioxari, P.~Doll{\'{a}}r, and R.~B. Girshick.
\newblock Mask {R-CNN}.
\newblock In {\em {ICCV}}, pages 2980--2988, 2017.

\bibitem{ResNet}
K.~He, X.~Zhang, S.~Ren, and J.~Sun.
\newblock Deep residual learning for image recognition.
\newblock In {\em {CVPR}}, pages 770--778, 2016.

\bibitem{RelGraph}
Y.~Hong, C.~R. Opazo, Y.~Qi, Q.~Wu, and S.~Gould.
\newblock Language and visual entity relationship graph for agent navigation.
\newblock In {\em {NeurIPS}}, pages 7685--7696, 2020.

\bibitem{RecVLN}
Y.~Hong, Q.~Wu, Y.~Qi, C.~R. Opazo, and S.~Gould.
\newblock {VLN} {BERT:} {A} recurrent vision-and-language {BERT} for
  navigation.
\newblock In {\em {CVPR}}, pages 1643--1653, 2021.

\bibitem{Transferable}
H.~Huang, V.~Jain, H.~Mehta, A.~Ku, G.~Magalh{\~{a}}es, J.~Baldridge, and
  E.~Ie.
\newblock Transferable representation learning in vision-and-language
  navigation.
\newblock In {\em {ICCV}}, pages 7403--7412, 2019.

\bibitem{R4R}
V.~Jain, G.~Magalh{\~{a}}es, A.~Ku, A.~Vaswani, E.~Ie, and J.~Baldridge.
\newblock Stay on the path: Instruction fidelity in vision-and-language
  navigation.
\newblock In {\em {ACL}}, pages 1862--1872, 2019.

\bibitem{ANewPath}
A.~Kamath, P.~Anderson, S.~Wang, J.~Y. Koh, A.~Ku, A.~Waters, Y.~Yang,
  J.~Baldridge, and Z.~Parekh.
\newblock A new path: Scaling vision-and-language navigation with synthetic
  instructions and imitation learning.
\newblock {\em CoRR}, abs/2210.03112, 2022.

\bibitem{Pathdreamer}
J.~Y. Koh, H.~Lee, Y.~Yang, J.~Baldridge, and P.~Anderson.
\newblock Pathdreamer: {A} world model for indoor navigation.
\newblock In {\em {ICCV}}, pages 14718--14728, 2021.

\bibitem{VG}
R.~Krishna, Y.~Zhu, O.~Groth, J.~Johnson, K.~Hata, J.~Kravitz, S.~Chen,
  Y.~Kalantidis, L.~Li, D.~A. Shamma, M.~S. Bernstein, and L.~Fei{-}Fei.
\newblock Visual genome: Connecting language and vision using crowdsourced
  dense image annotations.
\newblock {\em {IJCV}}, 123:32--73, 2017.

\bibitem{RxR}
A.~Ku, P.~Anderson, R.~Patel, E.~Ie, and J.~Baldridge.
\newblock Room-across-room: Multilingual vision-and-language navigation with
  dense spatiotemporal grounding.
\newblock In {\em {EMNLP}}, pages 4392--4412, 2020.

\bibitem{SEA}
C.~Kuo, C.~Ma, J.~Hoffman, and Z.~Kira.
\newblock Structure-encoding auxiliary tasks for improved visual representation
  in vision-and-language navigation.
\newblock In {\em {WACV}}, pages 1104--1113, 2023.

\bibitem{BLIP}
J.~Li, D.~Li, C.~Xiong, and S.~C.~H. Hoi.
\newblock {BLIP:} bootstrapping language-image pre-training for unified
  vision-language understanding and generation.
\newblock In {\em {ICML}}, pages 12888--12900.

\bibitem{Cross-Modal}
J.~Li, H.~Tan, and M.~Bansal.
\newblock Improving cross-modal alignment in vision language navigation via
  syntactic information.
\newblock In {\em {NAACL}}, pages 1041--1050, 2021.

\bibitem{SIVLN}
J.~Li, H.~Tan, and M.~Bansal.
\newblock Improving cross-modal alignment in vision language navigation via
  syntactic information.
\newblock In K.~Toutanova, A.~Rumshisky, L.~Zettlemoyer,
  D.~Hakkani{-}T{\"{u}}r, I.~Beltagy, S.~Bethard, R.~Cotterell, T.~Chakraborty,
  and Y.~Zhou, editors, {\em {NAACL-HLT}}, pages 1041--1050, 2021.

\bibitem{CLEAR}
J.~Li, H.~Tan, and M.~Bansal.
\newblock {CLEAR:} improving vision-language navigation with cross-lingual,
  environment-agnostic representations.
\newblock In {\em {NAACL}}, pages 633--649, 2022.

\bibitem{Envedit}
J.~Li, H.~Tan, and M.~Bansal.
\newblock Envedit: Environment editing for vision-and-language navigation.
\newblock In {\em {CVPR}}, pages 15386--15396, 2022.

\bibitem{liang2015automated}
C.~Liang, K.~Chee, Y.~Zou, H.~Zhu, A.~Causo, S.~Vidas, T.~Teng, I.~Chen,
  K.~Low, and C.~Cheah.
\newblock Automated robot picking system for e-commerce fulfillment warehouse
  application.
\newblock In {\em {IFToMM}}, 2015.

\bibitem{Prompt}
X.~Liang, F.~Zhu, L.~Li, H.~Xu, and X.~Liang.
\newblock Visual-language navigation pretraining via prompt-based environmental
  self-exploration.
\newblock In {\em {ACL}}, pages 4837--4851, 2022.

\bibitem{Contrastive-PI}
X.~Liang, F.~Zhu, Y.~Zhu, B.~Lin, B.~Wang, and X.~Liang.
\newblock Contrastive instruction-trajectory learning for vision-language
  navigation.
\newblock In {\em {AAAI}}, pages 1592--1600, 2022.

\bibitem{ADAPT}
B.~Lin, Y.~Zhu, Z.~Chen, X.~Liang, J.~Liu, and X.~Liang.
\newblock {ADAPT:} vision-language navigation with modality-aligned action
  prompts.
\newblock In {\em {CVPR}}, pages 15375--15385, 2022.

\bibitem{COCO}
T.~Lin, M.~Maire, S.~J. Belongie, J.~Hays, P.~Perona, D.~Ramanan,
  P.~Doll{\'{a}}r, and C.~L. Zitnick.
\newblock Microsoft {COCO:} common objects in context.
\newblock In {\em {ECCV}}, pages 740--755, 2014.

\bibitem{SIA}
X.~Lin, G.~Li, and Y.~Yu.
\newblock Scene-intuitive agent for remote embodied visual grounding.
\newblock In {\em {CVPR}}, pages 7036--7045, 2021.

\bibitem{ViLBERT}
J.~Lu, D.~Batra, D.~Parikh, and S.~Lee.
\newblock Vilbert: Pretraining task-agnostic visiolinguistic representations
  for vision-and-language tasks.
\newblock In {\em {NeurIPS}}, pages 13--23, 2019.

\bibitem{SMNA}
C.~Ma, J.~Lu, Z.~Wu, G.~AlRegib, Z.~Kira, R.~Socher, and C.~Xiong.
\newblock Self-monitoring navigation agent via auxiliary progress estimation.
\newblock In {\em {ICLR}}, 2019.

\bibitem{cat}
S.~Ma, Y.~Wang, Y.~Wei, J.~Fan, T.~H. Li, H.~Liu, and F.~Lv.
\newblock Cat: Localization and identification cascade detection transformer
  for open-world object detection.
\newblock In {\em {CVPR}}, pages 19681--19690, 2023.

\bibitem{Improving}
A.~Majumdar, A.~Shrivastava, S.~Lee, P.~Anderson, D.~Parikh, and D.~Batra.
\newblock Improving vision-and-language navigation with image-text pairs from
  the web.
\newblock In {\em {ECCV}}, pages 259--274, 2020.

\bibitem{VLNBert}
A.~Majumdar, A.~Shrivastava, S.~Lee, P.~Anderson, D.~Parikh, and D.~Batra.
\newblock Improving vision-and-language navigation with image-text pairs from
  the web.
\newblock In {\em {ECCV}}, pages 259--274, 2020.

\bibitem{SOAT}
A.~Moudgil, A.~Majumdar, H.~Agrawal, S.~Lee, and D.~Batra.
\newblock {SOAT:} {A} scene- and object-aware transformer for
  vision-and-language navigation.
\newblock In {\em {NeurIPS}}, pages 7357--7367, 2021.

\bibitem{EATA}
S.~Niu, J.~Wu, Y.~Zhang, Y.~Chen, S.~Zheng, P.~Zhao, and M.~Tan.
\newblock Efficient test-time model adaptation without forgetting.
\newblock In {\em {ICML}}, pages 16888--16905, 2022.

\bibitem{GPT4}
OpenAI.
\newblock {GPT-4} technical report.
\newblock 2023.

\bibitem{pytorch}
A.~Paszke, S.~Gross, F.~Massa, A.~Lerer, J.~Bradbury, G.~Chanan, T.~Killeen,
  Z.~Lin, N.~Gimelshein, L.~Antiga, A.~Desmaison, A.~K{\"{o}}pf, E.~Z. Yang,
  Z.~DeVito, M.~Raison, A.~Tejani, S.~Chilamkurthy, B.~Steiner, L.~Fang,
  J.~Bai, and S.~Chintala.
\newblock Pytorch: An imperative style, high-performance deep learning library.
\newblock In {\em {NeurIPS}}, pages 8024--8035, 2019.

\bibitem{REVERIE}
Y.~Qi, Q.~Wu, P.~Anderson, X.~Wang, W.~Y. Wang, C.~Shen, and A.~van~den Hengel.
\newblock {REVERIE:} remote embodied visual referring expression in real indoor
  environments.
\newblock In {\em {CVPR}}, pages 9979--9988, 2020.

\bibitem{HOP}
Y.~Qiao, Y.~Qi, Y.~Hong, Z.~Yu, P.~Wang, and Q.~Wu.
\newblock Hop: History-and-order aware pre-training for vision-and-language
  navigation.
\newblock In {\em {CVPR}}, pages 15418--15427, 2022.

\bibitem{CLIP}
A.~Radford, J.~W. Kim, C.~Hallacy, A.~Ramesh, G.~Goh, S.~Agarwal, G.~Sastry,
  A.~Askell, P.~Mishkin, J.~Clark, G.~Krueger, and I.~Sutskever.
\newblock Learning transferable visual models from natural language
  supervision.
\newblock In {\em {ICML}}, pages 8748--8763, 2021.

\bibitem{HMP3D}
S.~K. Ramakrishnan, A.~Gokaslan, E.~Wijmans, O.~Maksymets, A.~Clegg, J.~Turner,
  E.~Undersander, W.~Galuba, A.~Westbury, A.~X. Chang, M.~Savva, Y.~Zhao, and
  D.~Batra.
\newblock Habitat-matterport 3d dataset {(HM3D):} 1000 large-scale 3d
  environments for embodied {AI}.
\newblock In {\em NeurIPS Datasets and Benchmarks}, 2021.

\bibitem{Outdoor}
R.~Schumann and S.~Riezler.
\newblock Analyzing generalization of vision and language navigation to unseen
  outdoor areas.
\newblock In S.~Muresan, P.~Nakov, and A.~Villavicencio, editors, {\em {ACL}},
  pages 7519--7532, 2022.

\bibitem{One-Step}
C.~H. Song, J.~Kil, T.~Pan, B.~M. Sadler, W.~Chao, and Y.~Su.
\newblock One step at a time: Long-horizon vision-and-language navigation with
  milestones.
\newblock In {\em {CVPR}}, pages 15461--15470, 2022.

\bibitem{EnvDrop}
H.~Tan, L.~Yu, and M.~Bansal.
\newblock Learning to navigate unseen environments: Back translation with
  environmental dropout.
\newblock In J.~Burstein, C.~Doran, and T.~Solorio, editors, {\em {NAACL}},
  pages 2610--2621, 2019.

\bibitem{Transformer}
A.~Vaswani, N.~Shazeer, N.~Parmar, J.~Uszkoreit, L.~Jones, A.~N. Gomez,
  L.~Kaiser, and I.~Polosukhin.
\newblock Attention is all you need.
\newblock In {\em {NeurIPS}}, pages 5998--6008, 2017.

\bibitem{Less-is-More}
S.~Wang, C.~Montgomery, J.~Orbay, V.~Birodkar, A.~Faust, I.~Gur, N.~Jaques,
  A.~Waters, J.~Baldridge, and P.~Anderson.
\newblock Less is more: Generating grounded navigation instructions from
  landmarks.
\newblock In {\em {CVPR}}, pages 15407--15417, 2022.

\bibitem{RCMS}
X.~Wang, Q.~Huang, A.~Celikyilmaz, J.~Gao, D.~Shen, Y.~Wang, W.~Y. Wang, and
  L.~Zhang.
\newblock Reinforced cross-modal matching and self-supervised imitation
  learning for vision-language navigation.
\newblock In {\em {CVPR}}, pages 6629--6638, 2019.

\bibitem{RCM}
X.~Wang, Q.~Huang, A.~Celikyilmaz, J.~Gao, D.~Shen, Y.~Wang, W.~Y. Wang, and
  L.~Zhang.
\newblock Reinforced cross-modal matching and self-supervised imitation
  learning for vision-language navigation.
\newblock In {\em {CVPR}}, pages 6629--6638, 2019.

\bibitem{Gibson}
F.~Xia, A.~R. Zamir, Z.~He, A.~Sax, J.~Malik, and S.~Savarese.
\newblock Gibson env: Real-world perception for embodied agents.
\newblock In {\em {CVPR}}, pages 9068--9079, 2018.

\bibitem{GN}
L.~Xie, M.~Zhang, Y.~Li, W.~Qin, Y.~Yan, and E.~Yin.
\newblock Vision--language navigation with beam-constrained global
  normalization.
\newblock {\em {TNNLS}}, 2022.

\bibitem{HomeRoboticsApplication}
G.~A. Zachiotis, G.~Andrikopoulos, R.~Gornez, K.~Nakamura, and
  G.~Nikolakopoulos.
\newblock A survey on the application trends of home service robotics.
\newblock In {\em {ROBIO}}, pages 1999--2006, 2018.

\bibitem{TD}
Y.~Zheng and L.~Fan.
\newblock Moving object detection based on running average background and
  temporal difference.
\newblock In {\em {ISKE}}, pages 270--272, 2010.

\bibitem{Place_365}
B.~Zhou, {\`{A}}.~Lapedriza, A.~Khosla, A.~Oliva, and A.~Torralba.
\newblock Places: {A} 10 million image database for scene recognition.
\newblock {\em {TIPAMI}}, 40:1452--1464, 2018.

\bibitem{SOON}
F.~Zhu, X.~Liang, Y.~Zhu, Q.~Yu, X.~Chang, and X.~Liang.
\newblock {SOON:} scenario oriented object navigation with graph-based
  exploration.
\newblock In {\em {CVPR}}, pages 12689--12699, 2021.

\bibitem{AuxRN}
F.~Zhu, Y.~Zhu, X.~Chang, and X.~Liang.
\newblock Vision-language navigation with self-supervised auxiliary reasoning
  tasks.
\newblock In {\em {CVPR}}, pages 10009--10019, 2020.

\end{thebibliography}
}

\onecolumn 
\newpage
\appendix


\begin{center}
	{
		\LARGE{\textsc{Appendix}}
	}
\end{center}

In the supplementary, we provide more details of our method. We organize the supplementary as follows.

\begin{itemize}[leftmargin=*]
    \item In Section~\ref{sec:supp-collection}, we present more details on video collection.
    \item In Section~\ref{sec:supp-filter}, we present more details on frame filtering.
    \item In Section~\ref{sec:supp-trajctory}, we present more details on trajectory generation.
    \item In Section~\ref{sec:supp-instruction}, we present more details on instruction generation.
    \item In Section~\ref{sec:statistic-examples}, we present statistics and visualization examples of our YouTube-VLN dataset.
    \item In Section~\ref{sec:imple}, we provide more implementation details of experiments.
    \item In Section~\ref{sec:supp-layout_effectives}, we provide more details of the effectiveness of trajectory judgment task on layout reasoning ability.
    \item In Section~\ref{sec:supp-trans}, we provide the transferability results of our method.
    \item In Section~\ref{sec:supp-qualitive}, we present qualitative results of our method.
    \item In Section~\ref{sec:supp-future}, we discuss potential future research and social impact.
\end{itemize}

\section{More Details on YouTube Video Collection}
\label{sec:supp-collection}
We collect real estate tour videos from YouTube\footnote{\url{https://www.youtube.com}}. Specifically, we restrict the videos to real estate tour videos. It is unrealistic to play all categories of videos one by one to check whether they meet our requirements. On the contrary, we look for several well-known YouTubers whose playlists have been well-categorized for house tour videos, and each list has a consistent style. To find such YouTubers, we only spend less than an hour of mannual search time. In YouTube, each video has its corresponding video id (~\eg $C99YjG\_JBsg$\footnote{\url{https://www.youtube.com/watch?v=C99YjG_JBsg}}), we obtain all the house tour videos according to the video ids and regard the video ids as house ids.

\section{More Details on Frame Filtering}
\label{sec:supp-filter}
\begin{figure*}[h]
     \centering
     \begin{subfigure}{\textwidth}
         \centering
         \includegraphics[width=\textwidth]{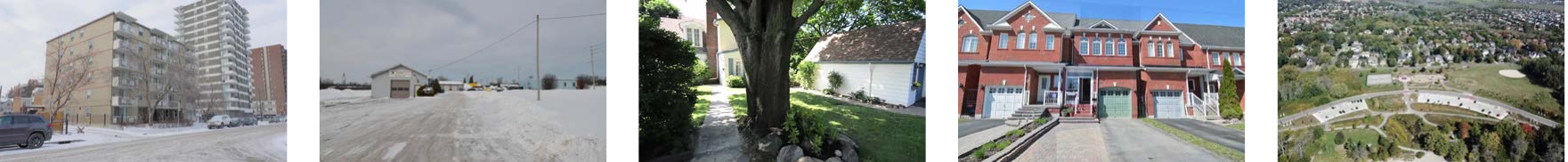}
         \caption{Frames with outdoor scenes.}
         \label{fig:outdoor_example}
     \end{subfigure}
     \hfill
     \begin{subfigure}{\textwidth}
         \centering
         \includegraphics[width=\textwidth]{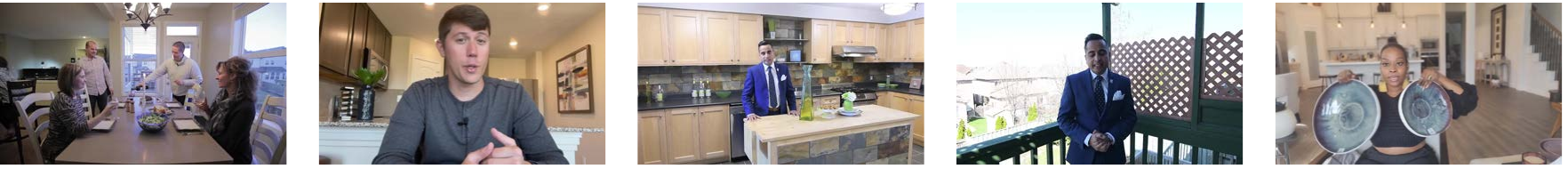}
         \caption{Frames with persons.}
         \label{fig:person_example}
     \end{subfigure}
     \begin{subfigure}{\textwidth}
         \centering
         \includegraphics[width=\textwidth]{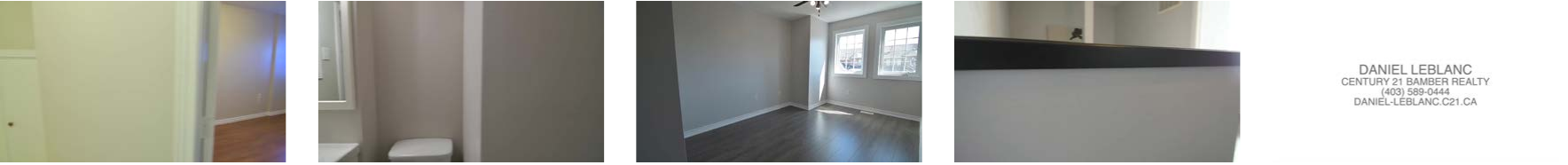}
         \caption{Frames without region features.}
         \label{fig:feature_example}
     \end{subfigure}
    \caption{Examples of filtered frames, including those with outdoor scenes~(a), persons~(b) or no region features~(c).}
    \label{fig:models}
\end{figure*}

Before generating a navigation trajectory, we first pre-process these videos by sparse sampling and using off-the-shelf image classifiers~\cite{Mask_RCNN, ResNet} to filter out redundant frames and noisy frames (\ie frames with persons or outdoor scenes). A video can be sampled at up to 60 frames per second. Generally, there is almost no obvious change between screens within 2s intervals in a video. Therefore, we sparsely sample the videos with 0.5 frames per second. Considering that in the downstream indoor VLN task, persons and outdoor images are not allowed to occur in the observations, we discard such noisy frames in the house tour videos. Specifically, we employ a Resnet~\cite{ResNet} model pre-trained on Place 365~\cite{Place_365} and Mask RCNN~\cite{Mask_RCNN} model pre-trained on COCO~\cite{COCO} to detect the outdoor images~(as shown in Figure~\ref{fig:outdoor_example}) and images with persons~(as shown in Figure~\ref{fig:person_example}), respectively. In addition, some images are filtered since they do not contain any objects and can not be extracted to region features (as shown in Figure~\ref{fig:feature_example}).

\begin{figure*}[htp]
     \centering
     \begin{subfigure}{0.52\textwidth}
         \centering
         \includegraphics[width=\textwidth]{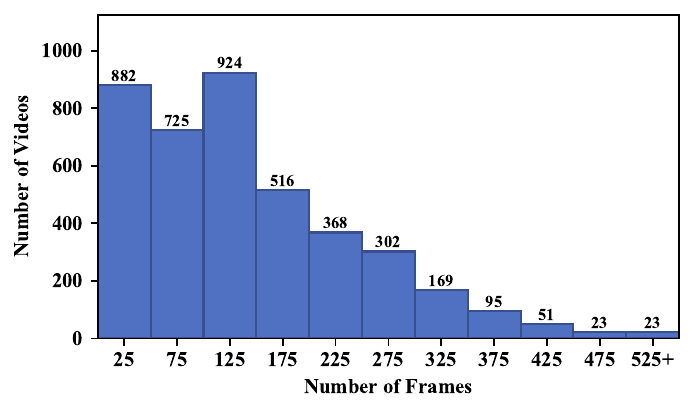}
         \caption{Distribution of the number of frames per video.}
         \label{fig:frames_video}
     \end{subfigure}
     \hspace{35pt}
     \begin{subfigure}{0.36\textwidth}
         \centering
         \includegraphics[width=\textwidth]{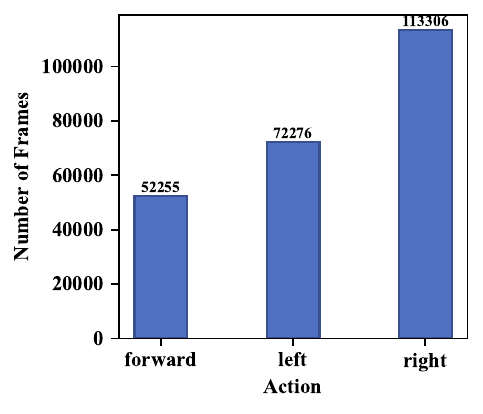}
         \caption{Distribution of actions.}
         \label{fig:actions}
     \end{subfigure}
     \hfill
     \begin{subfigure}{1.0\textwidth}
         \centering
         \includegraphics[width=\textwidth]{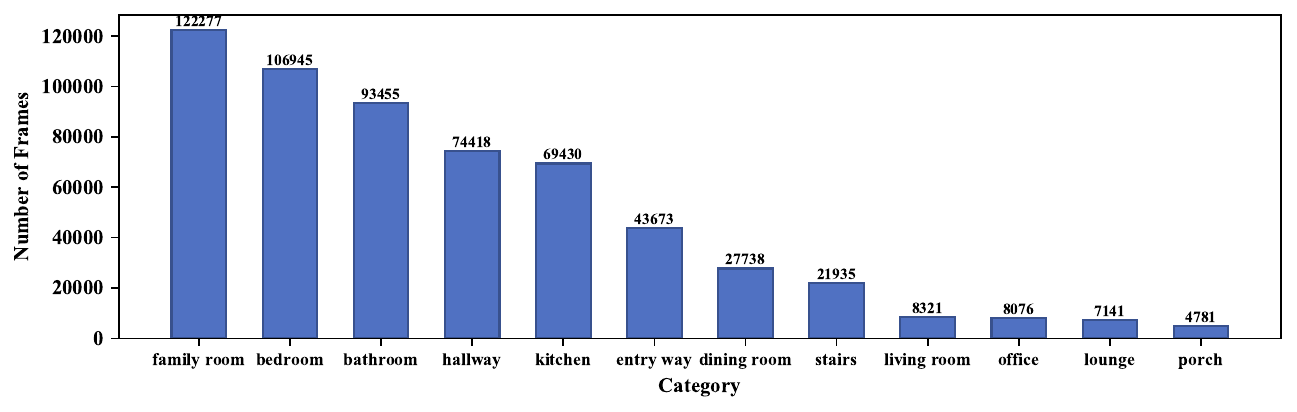}
         \caption{Distribution of predicted scene categories on YouTube frames.}
         \label{fig:room_types}
     \end{subfigure}
    \caption{Statistics of YouTube-VLN Dataset.}
    \label{fig:statistics}
\end{figure*}

\section{More Details on Trajectory Generation}
\label{sec:supp-trajctory}
To mimic the sequential navigation path in R2R, we typically choose $K \in [4,7]$ as the length of a trajectory. As mentioned in Section 3.1, a trajectory consists of room nodes and transition nodes. We randomly sample $R \in [2,7]$ room nodes in temporal order for a trajectory. Considering that 1) navigation is a continuous problem in both temporal dimension and spatial dimension, and 2) instruction does not necessarily describe all observations on a trajectory, the remaining $(K - R)$ nodes are filled with transition nodes. Each image is encoded into region features by Faster R-CNN bottom-up top-down attention~\cite{bottom-up} model pre-trained on Visual Genome~\cite{VG}. In order to approximate the panoramic visual context, we merge the region features from similar room types per image. The images we merged are from consecutive frames and in the same group, usually taken by the real estate agent around similar locations in order to better introduce the room.

\begin{figure*}[!t]
    \centering
    \includegraphics[width=1.0\linewidth]{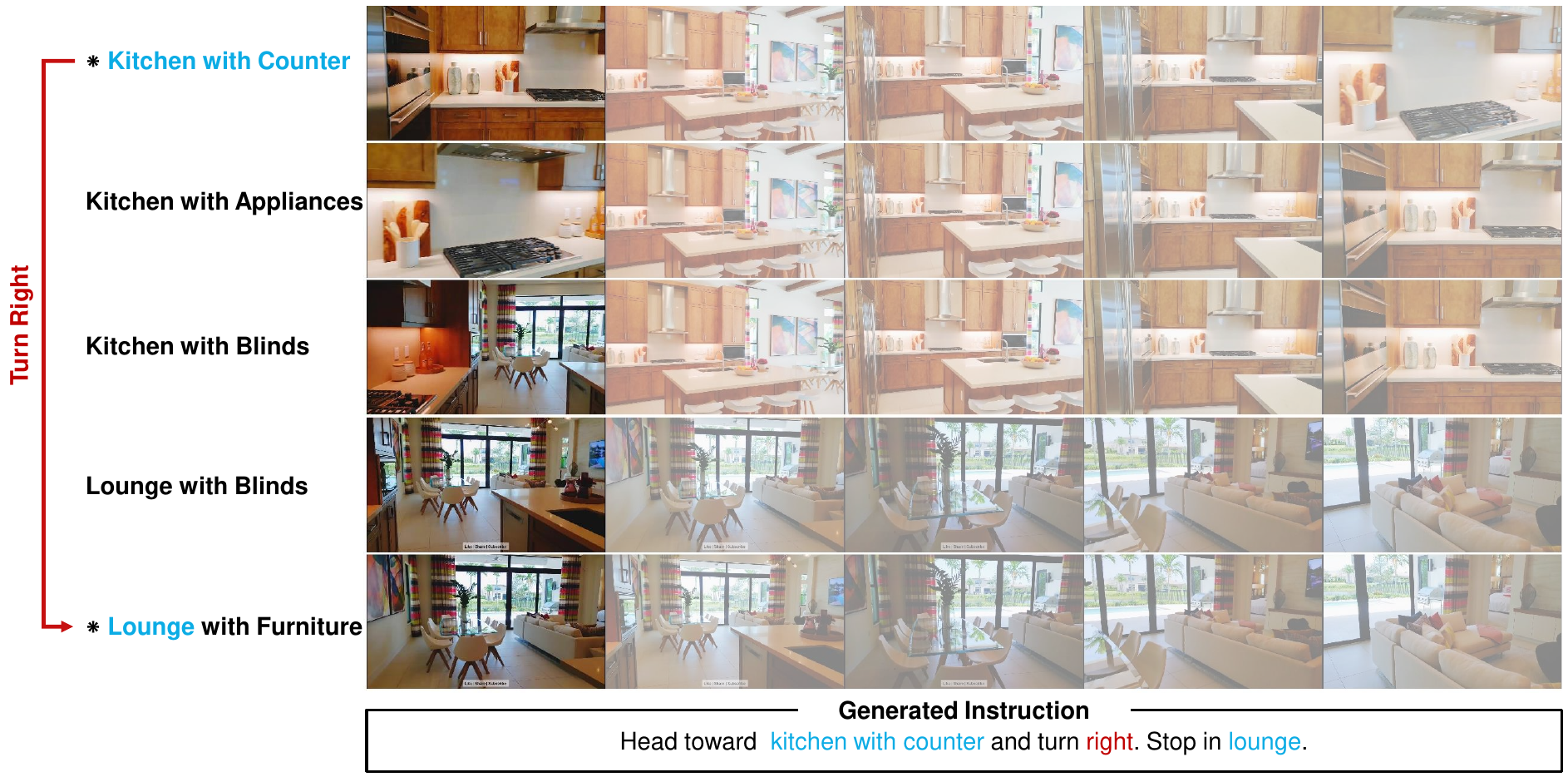}
    \caption{Examples of path-instruction pairs in the proposed YouTube-VLN dataset. The generated trajectory contains 2 room nodes (marked as *) and 3 transition nodes. The translucent images are the merged images.}
    \label{fig:YTb_1}
\end{figure*}

\begin{figure*}[!t]
    \centering
    \includegraphics[width=1.0\linewidth]{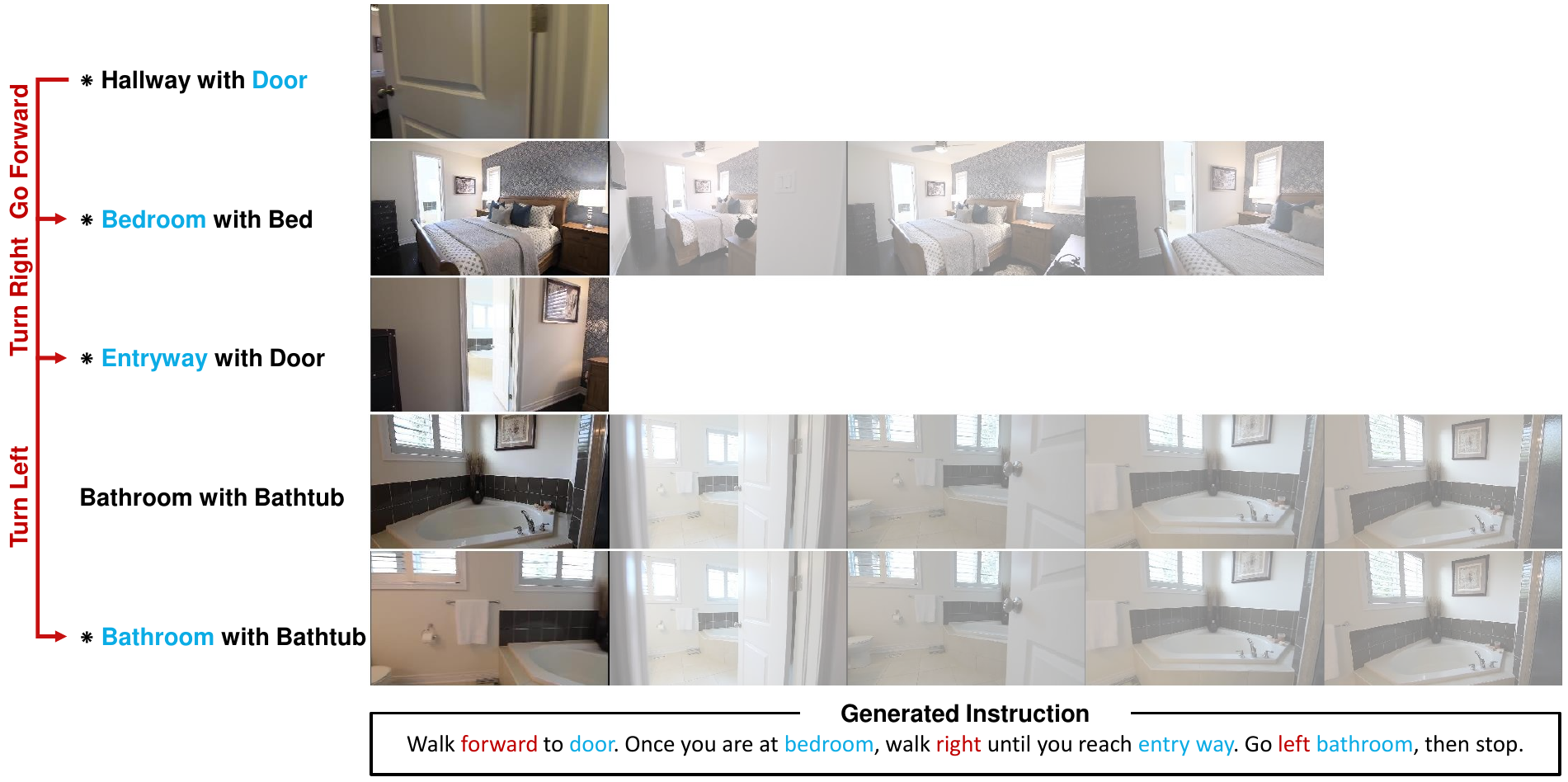}
    \caption{Examples of path-instruction pairs in the proposed YouTube-VLN dataset. The generated trajectory contains 4 room nodes (marked as *) and 1 transition nodes. The translucent images are the merged images.}
    \label{fig:YTb_2}
    \vspace{-3mm}
\end{figure*}

\begin{figure*}[!t]
    \centering
    \includegraphics[width=1.0\linewidth]{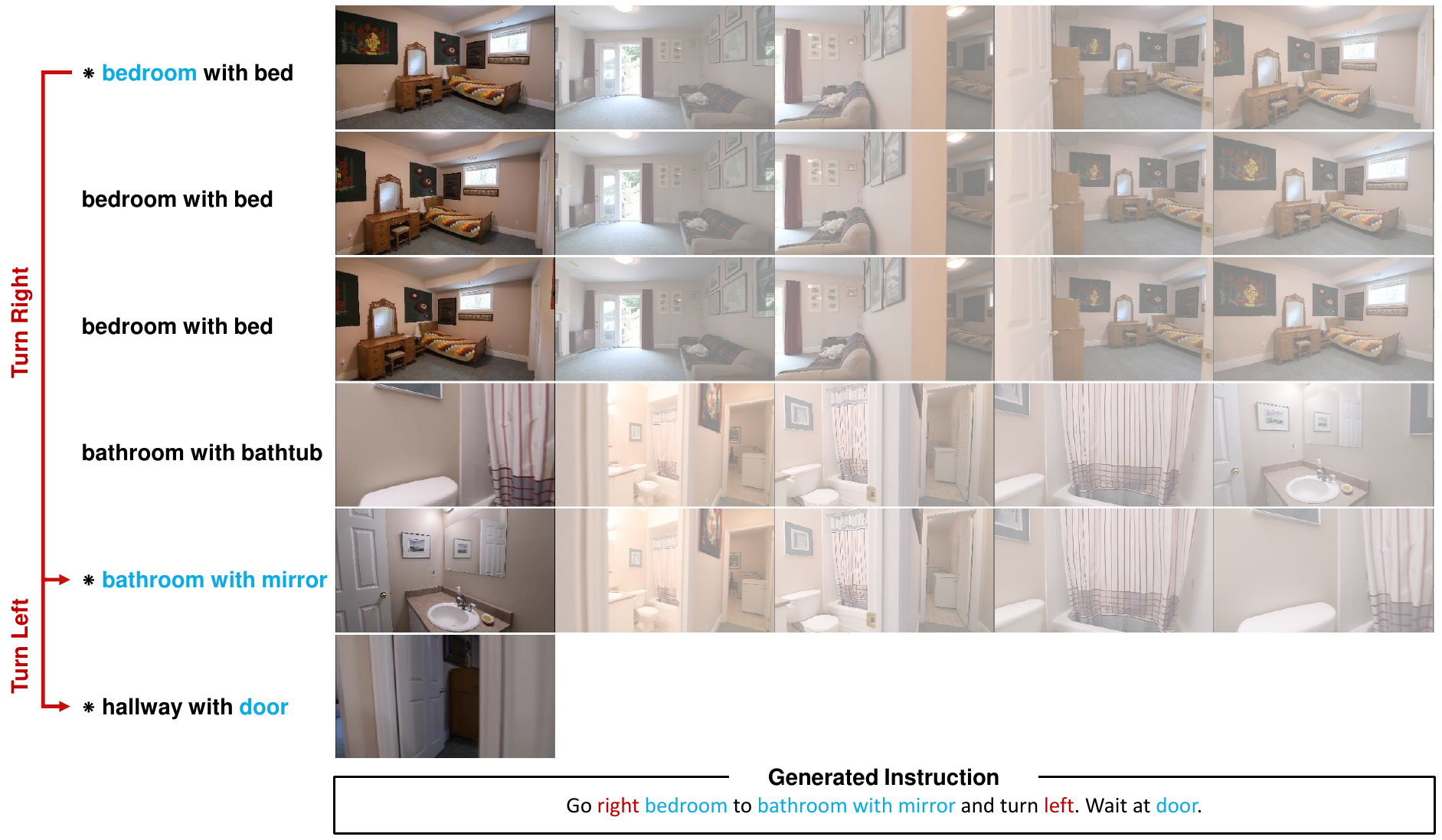}
    \caption{Examples of path-instruction pairs in the proposed YouTube-VLN dataset. The generated trajectory contains 3 room nodes (marked as *) and 3 transition nodes. The translucent images are the merged images.}
    \label{fig:YTb_3}
\end{figure*}

\section{More Details on Instruction Generation}
\label{sec:supp-instruction}
As for instruction generation, we first harvest 14,031 fill-in-the-blank templates from the R2R training set. Specifically, we extract the noun phrases and verb phrases for each human-annotated navigation instruction in the R2R training set. We then randomly select a template that has $R$ noun phrase blanks and $(R-1)$ verb phrase blanks for a trajectory with $R$ room nodes. We use CLIP~\cite{CLIP} model to caption the room nodes with a template ``[room] with [object]'' following~\cite{Prompt}, where the ``[room]'' and ``[object]'' represent the room category and object category of a room node, respectively. To better include the details of the rooms, we fill a noun blank with ``[room] with [object]'' or ``[room]'' or ``[object]''.  A CNN inverse action model~\cite{YTb} infers the transition action from one room node to another. Finally, we obtain $R$ captions and $(R-1)$ action words for a template. The captions are used to fill the noun phrase blanks in the template sequentially. For each noun phrase blank filled with the captions of one room node, we find its closest verb phrase blank and fill it with the action which is executed to reach the next room node. Our instruction generation strategy fills the verb phrase blanks with pseudo-labeled actions, providing a natural transition between two nodes to the created instruction.

\section{Statistics and Visualizations of YouTube-VLN Dataset}
\label{sec:statistic-examples}
In Figure~\ref{fig:statistics}, we show some key statistics about our YouTube-VLN dataset. We construct the YouTube-VLN from the collected 4078 videos. After filtering the noisy frames, we harvest 568K images in total. In Figure~\ref{fig:frames_video}, we present the number of frames per video via a histogram. It shows that most of the videos contain more than 25 effective frames, indicating that each video can provide sufficient image samples for an agent to learn and reason about this house. Figure~\ref{fig:room_types} presents the predicted room types of the frames. We used CLIP~\cite{CLIP} to categorize each frame into one of the 12 labeled room types in Matterport dataset~\cite{VLN}. It can be observed that most of the images in the proposed YouTube-VLN dataset cover the core part of a house (~\eg family room and bedroom). This enables the agent to learn the layout prior knowledge more efficiently. Moreover, these labels are further used for instruction generation and image merging. In addition, we also show the distribution of the pseudo-labeled actions in Figure~\ref{fig:actions}. Each action is the predicted native action from one room node to another, representing the direction that the agent should follow. As shown in the histogram, the pseudo-labeled actions are evenly distributed into three types of actions, endowing the agent to understand the actions efficiently. We also show some visualization examples of the generated path-instruction pairs in YouTube-VLN as in Figure~\ref{fig:YTb_1}, Figure~\ref{fig:YTb_2}, and Figure~\ref{fig:YTb_3}.


\section{More Implementation Details}
\label{sec:imple}
The model architecture details are shown in Figure~\ref{fig:models}. The meanings of each layer are as follows:

\begin{itemize}
    \item \textbf{Embed-Lang}: language token embeddings, which consist of word embeddings, position embeddings, and token type embeddings.
    \item \textbf{Embed-Vis}: vision token embeddings, which consist of visual feature embeddings and position embeddings of the current node.
    \item \textbf{Embed-Node}: node embeddings, which consist of visual feature embeddings, position embeddings, and navigation step embeddings of all nodes in the navigation graph.
    \item \textbf{Self-Att-Lang}: self-attention layers for language input.
    \item \textbf{Self-Att-Vis}: self-attention layers for vision input.
    \item \textbf{Cross-Att-Vis}: cross-modal attention layers for vision branch.
    \item \textbf{Cross-Att-Lang}: cross-modal attention layers for language branch.
    \item \textbf{FFN}: feed-forward network, which consists of two linear layers.
\end{itemize}

\begin{figure*}[!t]
     \centering
     \begin{subfigure}{1.0\textwidth}
         \centering
         \includegraphics[width=\textwidth]{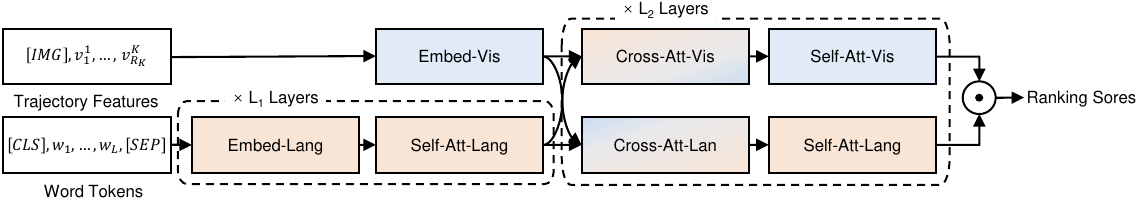}
         \caption{Adapting Lily agent to the discriminative setting (same as Airbert~\cite{Airbert}).}
         \label{fig:vilbert_disc_model}
     \end{subfigure}\vspace{3mm}
     \hfill
     \begin{subfigure}{1.0\textwidth}
         \centering
         \includegraphics[width=\textwidth]{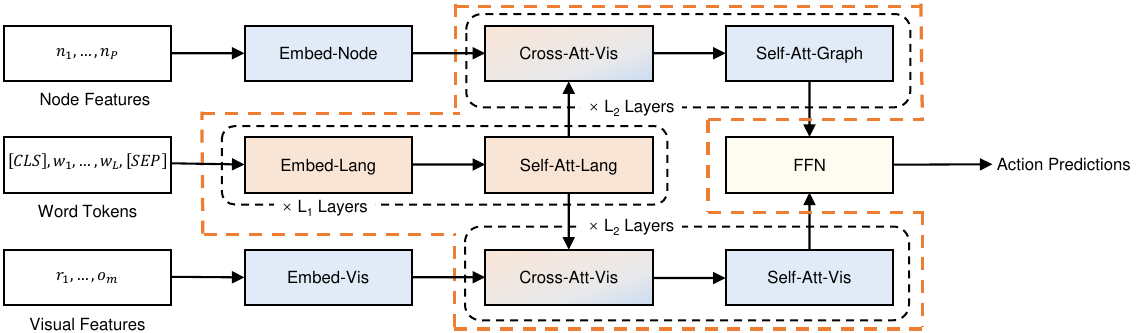}
         \caption{Adapting Lily agent to the generative setting (same as DUET~\cite{DUET}).}
         \label{fig:vilbert_gen_model}
     \end{subfigure}
    \caption{The adapted model in both discriminative and generative settings for downstream VLN tasks.}
    \label{fig:models}
\end{figure*}

\subsection{Pre-training Details}
As described in Section~4.1, we adopt a ViLBERT-like architecture as the same as Airbert~\cite{Airbert}. The model architecture of pre-training is shown in Figure~\ref{fig:vilbert_disc_model}. For a fair comparison, we set both $L_1$ and $L_2$ to 6, consistent with Airbert in the discriminative setting. For generative adaption, the number of layers $L_1$ is equal to 9 and $L_2$ is equal to 5 for a fair comparison with DUET~\cite{DUET}. For both settings, we distribute training over 4 NVIDIA 3090 GPUs~(24GB each) for 500 epochs to convergence. The batch size is 8 (2 for each GPU) and the learning rate is $2 \times 10^{-5}$. We randomly selected 95\% videos per epoch as the training set and 5\% videos as the test set.

\subsection{Fine-tuning Details}

\paragraph{Discriminative Setting.}
In our experiments, discriminative evaluation is conducted on R2R dataset. In this setting, VLN is formulated as a path selection problem. As shown in Figure~\ref{fig:vilbert_disc_model}, the model architecture is the same as the pre-trained model whose classifier used in the path ranking pretext task can be directly for path selection. We follow a two-stage fine-tuning as Airbert, which fine-tunes the agent with MLM task and MVM task in stage one and PR task in stage two. In stage one, the batch size is 12 on 4 VIDIA 3090 GPUs~(24GB each) and the learning rate is $4 \times 10^{-5}$. In stage two, we set the batch size as 16 on 8 NVIDIA 3090 GPUs~(24GB each) and the learning rate as $1 \times 10^{-5}$. The agent is fine-tuned for 30 epochs in both stages. The visual features are also encoded by a bottom-up top-down attention~\cite{bottom-up} model. We select the model checkpoint with the highest success rate on the val unseen validation split for the test set evaluation and leaderboard submission.

\vspace{-3mm}
\paragraph{Generative Setting.}
In the generative setting, the agent needs to predict actions sequentially in order to reach the goal (R2R) or find the object (REVERIE). We adopt DUET~\cite{DUET} as the architecture for fine-tuning as it is the state-of-the-art model. As illustrated in Figure~\ref{fig:vilbert_gen_model}, DUET is a three-stream  architecture which is fed with $P$ node features $\left\{n_i\right\}_{i=1}^P$, word tokens $\{[\texttt {CLS}],\left\{w_i\right\}_{i=1}^L, [\texttt {SEP}]\}$ and the current panorama encoding with image features $\left\{r_i\right\}_{i=1}^n$ together with object features $\left\{o_i\right\}_{i=1}^m$. The BERT-like architecture is used to determine which node should the agent go to or what the object goal id is. We initialize the language stream and the cross-modal streams using the corresponding modules in our pre-trained model, highlighted as the orange dotted line in Figure~\ref{fig:vilbert_gen_model}. The other settings remain the same as DUET for a fair comparison.

\begin{figure*}[t]
     \centering
     \begin{subfigure}{0.45\textwidth}
         \centering
         \includegraphics[width=\textwidth]{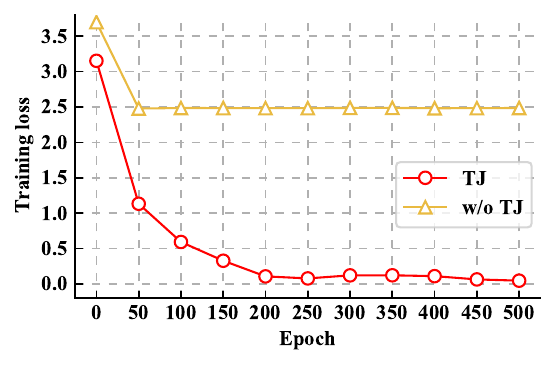}
     \end{subfigure}
     \hfill
     \begin{subfigure}{0.45\textwidth}
         \centering
         \includegraphics[width=\textwidth]{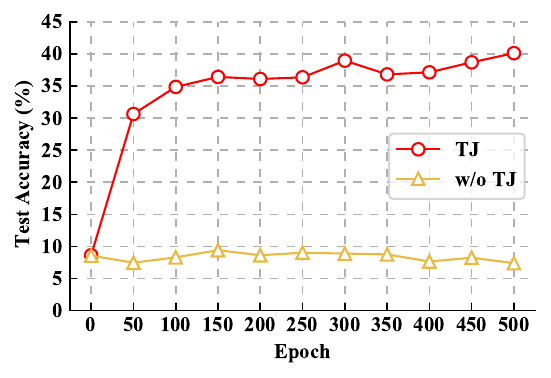}
     \end{subfigure}
    \caption{Evolution of training cross-entropy (CE) loss and test accuracy \wrt training epochs on R2R dataset.}
    \label{fig:layout}
\end{figure*}

\section{Effectiveness of Trajectory Judgment Task on Layout Reasoning Ability}
\label{sec:supp-layout_effectives}
To evaluate the effectiveness of the proposed trajectory judgment task, we conduct an experiment to evaluate whether the agent trained with this task is able to figure out the direction of an unexplored room based on current observation. Since the common room layout knowledge is required for figuring out this question, the accuracy of this question reveals the layout reasoning ability of the agent.

We conduct this experiment on the R2R~\cite{VLN} dataset, which provides a navigation graph for each environment. Specifically, we randomly initialize an agent on a navigation node. Then, we sample another node from the candidate node list of the current node and use the CLIP~\cite{CLIP} model to identify the room type of this node. Given this room type and the panorama visual feature of the current node as text input and visual input, respectively, the agent is asked to predict the relative orientation of the node for that room type. We define the angle between the matching node of the room type and the current orientation as $s \in [-180^\circ, 180^\circ]$. We then divide $[-180^\circ, 180^\circ]$ uniformly into twelve intervals and compute the interval that $s$ belongs to. This task thus becomes a twelve-category problem and is optimized by minimizing the cross-entropy loss:
\vspace{-2mm}
\begin{equation}
L_{C E}=-\sum_{i=1}^{12} y_i \ln \hat{y_i}
\vspace{-2mm}
\end{equation}
where $\hat{y_i}=\frac{e^{-x_i}}{\sum_{k=1}^{12} e^{-x_k}}$ represents the predicted probability of $s$ belonging to $i^{th}$ interval, $x_i$ represents $i^{th}$ output logit of the model and $y_i \in\{0,1\}$ indicates whether $s$ belongs to $i^{th}$ interval.
We train two agents for 500 epochs, one with the proposed trajectory judgment task and one without it. 

As shown in Figure \ref{fig:layout}, the training cross-entropy loss almost does not decrease without being pre-trained with the trajectory judgment task. This indicates that this variant has not learned the layout reasoning ability at all. Pre-trained with the trajectory judgment task, the agent model drops the training cross-entropy loss rapidly and the test accuracy increases stably. Finally, the highest accuracy of the variant pre-trained with the trajectory judgment task reaches around 40\%, while the other variant is 10\% (nearly equal to $ \frac{1}{12}$).  These results verify that the trajectory judgment task facilitates learning layout reasoning ability, achieving substantial improvement.

\begin{figure*}[!t]
    \centering
    \includegraphics[width=1.0\linewidth]{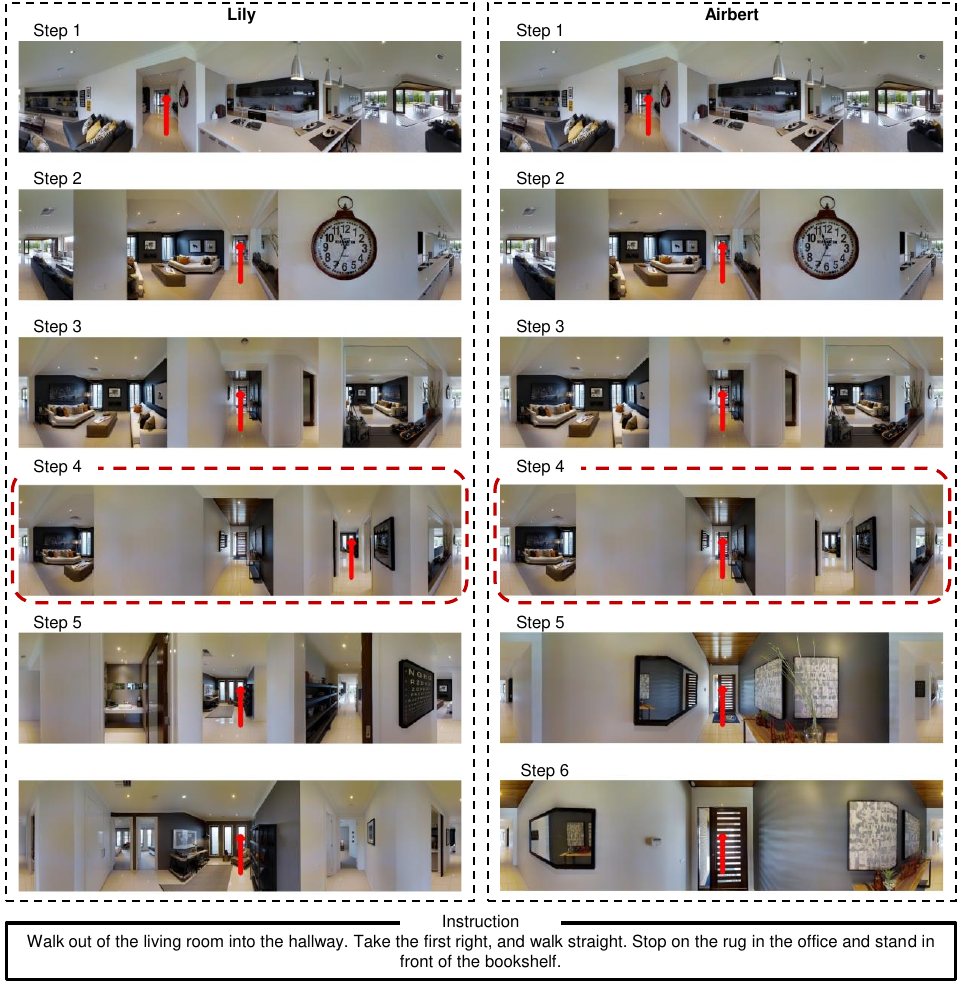}
    \caption{Visualisation of a trajectory where we compare the performance of our Lily agent with Airbert. The centre of each panorama is the heading direction of the agent at the corresponding time step. Red arrows indicate the predicted actions in each time step. Our Lily agent successfully leverages the layout prior knowledge to find the office.}
    \label{fig:layout_win}
\end{figure*}

\begin{figure*}[!t]
    \centering
    \includegraphics[width=1.0\linewidth]{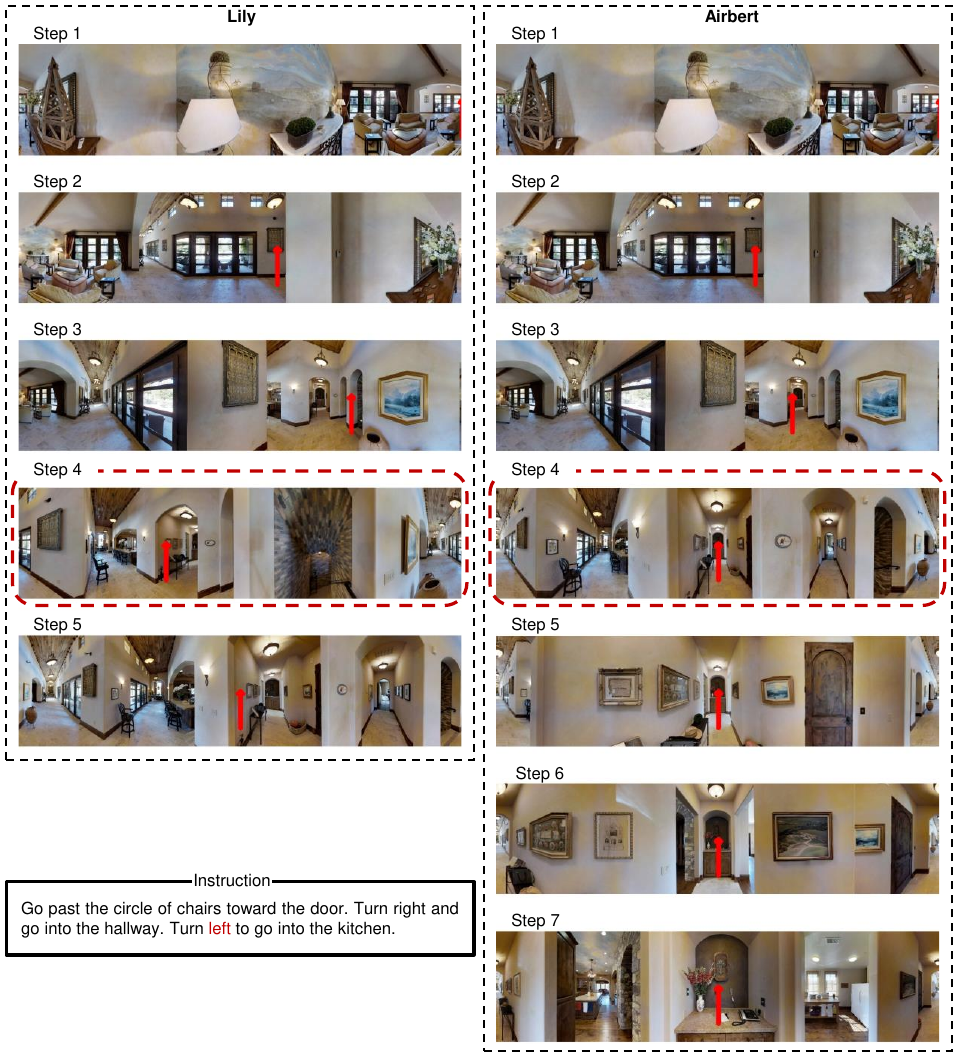}
    \caption{Visualisation of a trajectory where we compare the performance of our Lily agent with Airbert. The centre of each panorama is the heading direction of the agent at the corresponding time step. Red arrows indicate the predicted actions in each time step. Our Lily agent correctly understands the action in the instruction and executes it.}
    \label{fig:action_win}
\end{figure*}

\section{Transferability Results to Other VLN Benchmarks}
\label{sec:supp-trans}
To better confirm the effectiveness of the our method, we evaluate the transferability of Lily on the other three VLN benchmarks, \ie RxR~\cite{RxR}, R4R~\cite{R4R} and SOON~\cite{SOON}. Specifically, we transfer the model trained on R2R to RxR and R4R and the model trained on REVERIE to SOON without fine-tuning. In Table~\ref{tab:other}, most of the results share the same trend as the results on the R2R and REVERIE dataset, \ie consistently surpassing the SOTA with large margins on val unseen split. These results demonstrate that our method can generalize well to different domains with varying complexity.

\begin{table}[h]
\centering
\resizebox{0.5\linewidth}{!}{
\begin{tabular}{llcclcclcl}
\topline
\multirow{2}{*}{Methods} &  & \multicolumn{2}{c}{RxR} &  & \multicolumn{2}{c}{R4R} &  & \multicolumn{2}{c}{SOON}       \\ \cmidrule{3-4} \cmidrule{6-7} \cmidrule{9-10} 
                         &  & SR         & SPL        &  & SR         & SPL        &  & SR    & \multicolumn{1}{c}{SPL} \\ \midline
DUET~\cite{DUET}                    &  & 23.05      & 18.05       &  & 16.01       &13.20        &  & 2.83    & 2.09                    \\ Lily~(ours)                     &  & \textbf{27.20}      & \textbf{20.51}      &  & \textbf{20.76}      & \textbf{17.34}      &  & \textbf{5.72}       & \textbf{4.10}                    \\
\bottomline
\end{tabular}
}
\caption{Transfer results on RxR, R4R and SOON under val unseen split.}
\label{tab:other}
\end{table}
\vspace{-5mm}

\section{Qualitative Results}
\label{sec:supp-qualitive}
We also present some visualization examples of our Lily agent and the state-of-the-art agent Airbert on the R2R dataset. As shown in Figure~\ref{fig:layout_win}, given the instruction that asks the agent to go to a dining room, our Lily agent is able to arrive at the office more quickly than Airbert. We speculate that our Lily agent can be aware of the layout knowledge that an office is usually located in a room on either side of a hallway. Hence, our Lily agent goes straight to the hallway (the red arrow in step 4), while Airbert goes to an entryway connecting the door to the outside(the red arrow in step 4).
Our method also improves the understanding of the actions in an instruction. In Figure~\ref{fig:action_win}, our Lily agent can easily understand the instruction which asks it to  turn left~(the red arrow in step 4). However, when pre-trained with incorrect actions, Airbert feels confused about the action words and does not execute the ``Turn left'' in the instruction, and keeps going forward~(the red arrow in step 4), eventually not finding the kitchen.

\section{Potential Future Research and Social Impact}
\label{sec:supp-future}
Our method is still restricted to graph-based environments. In a real-world application, we may expect the agent actuates the action continuously. This requires us to build more continuous navigation trajectories for the pre-training dataset. Besides, with more powerful vision and language foundation models~\cite{BLIP, GPT4, cat}, the models used to construct the proposed dataset can be further improved as more precise and open-world. We can also increase the data diversity by adding richer video and instruction templates. 

In the future, agents can be able to actively learn some helpful skills by watching videos like us humans and then assist people with their jobs~(\eg delivering and cleaning), thereby reducing high training costs. However, data security can be an important issue. For some videos that humans keep secret or do not want agents to see, countermeasures should be taken to prevent agents from accessing such videos, otherwise, it may affect human survival one day.

\end{document}